\documentclass{article}
\usepackage[utf8]{inputenc}
\usepackage{geometry}[margin=0.5in]
\usepackage{cite}
\usepackage{amsmath,amssymb,amsfonts, amsthm}
\usepackage{algorithmic}
\usepackage{graphicx}
\usepackage{textcomp}

\newtheorem{theorem}{Theorem}

\newtheorem{problem}{Problem}

\DeclareMathOperator*{\expect}{\mathbb{E}}

\begin{document}

\title{Regret-optimal Estimation and Control}
\author{Gautam Goel and Babak Hassibi
\thanks{Gautam Goel is with the Department of Computing and Mathematical Sciences at Caltech (e-mail: ggoel@caltech.edu). }
\thanks{Babak Hassibi is with the Department of Electrical Engineering at Caltech (e-mail: bhassibi@caltech.edu).}}
\date{}

\maketitle

\begin{abstract}
We consider estimation and control in linear time-varying dynamical systems from the perspective of regret minimization. Unlike most prior work in this area, we focus on the problem of designing causal estimators and controllers which compete against a clairvoyant noncausal policy, instead of the best policy selected in hindsight from some fixed parametric class.  We show that the regret-optimal estimator and regret-optimal controller can be derived in state-space form using operator-theoretic techniques from robust control and present tight, data-dependent bounds on the regret incurred by our algorithms in terms of the energy of the disturbances. Our results can be viewed as extending traditional robust estimation and control, which focuses on minimizing worst-case cost, to minimizing worst-case regret. We propose regret-optimal analogs of Model-Predictive Control (MPC) and the Extended Kalman Filter (EKF) for systems with nonlinear dynamics and present numerical experiments which show that our regret-optimal algorithms can significantly outperform standard approaches to estimation and control.
\end{abstract}

% \begin{abstract}
% We consider estimation and control in linear time-varying dynamical systems from the perspective of regret minimization. Unlike most prior work in this area, we focus on the problem of designing causal estimators and controllers which compete against a clairvoyant noncausal policy, instead of the best policy selected in hindsight from some fixed parametric class.  This formulation is attractive when the environment changes over time and no fixed policy achieves good performance over the entire time horizon. We show that the regret-optimal estimator and regret-optimal controller can be derived in state-space form using operator-theoretic techniques from robust control and present tight, data-dependent bounds on the regret incurred by our algorithms in terms of the energy of the disturbances. We benchmark our regret-optimal controller in a nonlinear system and show that it can significantly outperform standard approaches to control.
% \end{abstract}

\section{Introduction}
The central question in control theory is how to regulate the behavior of an evolving system with state $x$ that is perturbed by a  disturbance $w$ by dynamically adjusting a control action $u$. Traditionally, this question has been studied in two distinct settings: in the $H_2$ setting, we assume that the disturbance $w$ is generated by a stochastic process and seek to select the control $u$ so as to minimize the \textit{expected} control cost, whereas in the $H_{\infty}$ setting we assume the noise is selected adversarially and instead seek to minimize the \textit{worst-case} control cost.

Both $H_2$ and $H_{\infty}$ controllers suffer from an obvious drawback: they are designed with respect to a specific class of disturbances, and if the true disturbances fall outside of this class, the performance of the controller may be poor. Indeed, the loss in performance can be arbitrarily large if the disturbances are carefully chosen \cite{doyle1978guaranteed}.

This observation naturally motivates the design of \textit{adaptive} controllers, which dynamically adjust their control strategy as they sequentially observe the disturbances instead of blindly following a prescribed strategy. This problem has attracted much recent attention in the online learning community (e.g. \cite{agarwal2019online, hazan2020nonstochastic, foster2020logarithmic}), mostly from the perspective of regret minimization. In this framework, the online controller is designed to minimize regret against the best controller selected in hindsight out of some class of controllers; the comparator class is often taken to be the class of state-feedback controllers or the class of disturbance-action controllers introduced in \cite{agarwal2019online}. The resulting controllers are adaptive in the sense that they seek to minimize cost irrespective of how the disturbances are generated.

In this paper, we take a somewhat different approach to the design of adaptive controllers. Instead of designing an online controller to minimize regret against the best controller selected in hindsight from some specific class, we instead focus on designing an online controller which minimizes regret against a clairvoyant noncausal controller, which knows the full sequence of disturbances in advance. The cost incurred by this noncausal controller is a lower bound on the cost incurred by any controller, since the noncausal controller selects the globally optimal sequence of control actions with full knowledge of the disturbances. This formulation of regret minimization is much more general: instead of imposing \textit{a priori} some finite-dimensional, parametric structure on the controller we learn (e.g. state-feedback policies, disturbance action policies, etc), which may or may not be appropriate for the given control task, we compete with the globally optimal clairvoyant controller, with no artificial constraints. We ask: how closely can  an online controller approximate the performance of a clairvoyant, noncausal controller? We design a new regret-optimal controller which approximates the performance of the optimal clairvoyant controller as closely as possible, and bound the worst-case difference in control costs between these two controllers (the regret) in terms of the energy of the disturbances. %Secondly, the controllers we obtain are more robust to changes in the environment. Consider, for example, a scenario in which an online controller is trying to compete with the best state-feedback controller selected in hindsight, and suppose the disturbances alternate between being generated by a stochastic process and being generated adversarially. When the disturbances are stochastic, an optimistic controller (such as the $H_2$ controller) will perform well; conversely, when the disturbances are adversarial, a more conservative, pessimistic controller (such as an $H_{\infty}$ controller) will perform well. No single state-feedback controller will perform well over the entire time horizon, and hence any online algorithm which tries to learn the best static state-feedback controller will incur high cumulative cost. A controller which minimizes regret against the noncausal controller, however, is not constrained to converge to any fixed controller, and hence can potentially outperform standard regret-minimizing control algorithms when the environment is dynamic.

We also consider online estimation (filtering) in dynamical systems from the perspective of regret minimization. Given a choice between a smoothed estimator and a filter, it is natural to prefer the smoothed estimator, since the smoothed estimator has access to noncausal measurements and can hence potentially outperform the filter. However, in many real-world settings we have only causal access to measurements and must generate estimates online using a filter. We ask: how closely can a filter approximate the performance of a smoothed estimator? We design a new regret-optimal filter which approximates the optimal smoothed estimator as closely as possible, and bound the worst-case difference in estimation error between these two estimators (the regret) in terms of the energy of the disturbances.

Our approach to regret minimization in estimation and control is similar in outlook to a series of works in online learning (e.g. \cite{herbster1998tracking, hazan2009efficient, jadbabaie2015online,  goel2019beyond, zinkevich2003online}), which seek to design online learning algorithms which compete with a globally optimal, dynamic sequence of actions instead of the best fixed action selected in hindsight from some class (e.g. the best fixed arm in the Multi-Armed Bandit problem). This notion of regret is called ``dynamic regret" and is natural when the reward-generating process encountered by the online algorithm varies over time. Unlike these prior works, we consider online optimization in settings with \textit{dynamics}. This setting is considerably more challenging to analyze through the lens of regret, because the dynamics serve to couple costs across rounds; the estimates or control actions generated by a learning algorithm in one round affect the costs incurred in all subsequent rounds, making counterfactual analysis difficult. 

%It is well-known that dynamic regret can scale linearly in the number of rounds in the worst case (\cite{jadbabaie2015online}). For this reason, dynamic regret bounds are usually stated in a ``data-dependent" manner, where the regret is bounded in terms of the variation or energy of the input sequence. For example, \cite{besbes2015non, wei2018more} and \cite{bubeck2019improved} give dynamic regret bounds for bandit optimization in terms of the pathlength $\sum_{t=1}^{T} \| \ell_t - \ell_{t-1}\|$, which measures the variation in the sequence of losses $\ell_1, \ldots \ell_T$; \cite{goel2019online} also give a dynamic regret bound in terms of pathlength in the context of online convex optimization with switching costs.. In recent work, \cite{baby2019online} and \cite{raj2020non} give dynamic regret bounds for online regression in terms of the total variation of the underlying sequence being estimated. In this paper, we derive the online control policy whose dynamic regret has optimal dependence on the energy of the disturbance $w$, namely $\sum_{t=0}^{T-1} \|w_t\|_2^2$; we call this policy the \textit{regret-optimal} control policy.

\subsection{Contributions of this paper}
In this paper, we consider finite-horizon estimation and control in linear dynamical systems from the perspective of regret minimization. We obtain two main results. First, we derive a new causal estimator (a filter) which minimizes regret against a noncausal estimator which receives all the full sequence of observations at once (i.e. the optimal smoothing estimator). This filter is drop-in replacement for standard filters such as the Kalman filter and the $H_{\infty}$ filter. We present a state-space model of the regret-optimal filter by constructing a new linear system such that the $H_{\infty}$ filter in the new system is the regret-optimal filter in the original system. The new system has $3n$ states, where $n$ is the number of states in the original system. Second, we derive a new causal controller which minimizes regret against a noncausal controller which receives the full sequence of disturbances in advance; this controller can be used in place of standard $H_2$ and $H_{\infty}$ controllers.  Given an $n$-dimensional linear control system and a corresponding cost functional, we show how to construct a $2n$-dimensional linear system and a new cost functional such that the $H_{\infty}$-optimal controller in the new system minimizes regret against the noncausal controller in the original system. Our results easily extend to settings where the controller has access to predictions of the next $k$ disturbances, or only affects the system dynamics after a delay of $k$ timesteps (Section \ref{predictions-delay-sec})
We also present tight data-dependent regret bounds for both the regret-optimal filter and the regret-optimal controller in terms of the energy of the disturbances. Our results can be viewed as extending traditional $H_{\infty}$ estimation and control, which focuses on minimizing worst-case \textit{cost}, to minimizing worst-case \textit{regret}.

We next consider nonlinear systems and describe how to obtain regret-optimal filters and controllers via iterative linearization; these schemes can be viewed as regret-optimal analogs of the classic Model Predictive Control (MPC) and Extended Kalman Filter (EKF) algorithms. We present numerical experiments which show that our regret-optimal algorithms can significantly outperform standard $H_2$ and $H_{\infty}$ algorithms across a wide variety of input disturbances. All of the algorithms we obtain are computationally efficient, and run in time linear in the time horizon.

\subsection{Related work}
Regret minimization in control has has attracted much recent attention across several distinct settings. A series of papers (e.g. \cite{abbasi2011regret, dean2018regret, cohen2019learning, lale2020regret}) consider a model where a linear system with unknown dynamics is perturbed by stochastic disturbances; an online learner picks control actions with the goal of minimizing regret against the optimal stabilizing controller. In the ``non-stochastic control" setting proposed in \cite{hazan2020nonstochastic}, the learner knows the dynamics, but the disturbance may be generated adversarially; the controller seeks to minimize regret against the class of disturbance-action policies. An $O(T^{2/3})$ regret bound was given in \cite{hazan2020nonstochastic}; this was improved to $O(T^{1/2})$ in \cite{agarwal2019online} and  $O(\log{T})$ in \cite{foster2020logarithmic}.

We emphasize that all of these works focus on minimizing regret against a fixed controller from some parametric class of control policies (policy regret), while we focus on competing against the clairvoyant noncausal controller; similar problems were also studied in, e.g. \cite{goel2017thinking, goel2019online, shi2020online}. Regret minimization was studied in LTI, infinite horizon estimation \cite{pmlr-v130-sabag21a} and control \cite{sabagcontrol}, and in finite-horizon measurement-feedback control in \cite{goelmeasurementfeedback}. Gradient-based algorithms with low dynamic regret against the class of disturbance-action policies were obtained in \cite{gradu2020adaptive, zhao2021non}.

\section{Preliminaries}
We consider estimation and control in linear time-varying dynamical systems over a finite-horizon $t = 0, \ldots T-1$. An estimator is \textit{causal} if its estimate at time $t$ depends only on the observations $y_0, \ldots, y_t$; otherwise we say the estimator is \textit{noncausal}. Causal estimation is often called \textit{filtering}; noncausal estimation is often called \textit{smoothing}. Similarly, a controller is causal if the control action it selects at time $t$ depends only on the disturbances $w_0, \ldots, w_{t-1}$; otherwise the controller is \textit{noncausal}. A strictly causal estimator or controller is defined analogously, except that its output at time $t$ depends only on the inputs up to time $t-1$. For simplicity we focus on causal estimation and control, but we emphasize that our results easily extend to the strictly causal setting as well; we describe how to adjust our proofs in the Appendix. We often think of estimators or controllers as being represented by \textit{linear operators} mapping disturbances $w$ to control actions $u$; in this framework, a (strictly) causal estimator or controller is precisely one whose associated operator is (strictly) lower-triangular.

In the estimation setting, a system with state $x_t \in \mathbb{R}^n$ evolves according to the dynamics 
\begin{equation} \label{estimation-dynamics-model}
x_{t+1} = A_t x_t + B_t u_t,
\end{equation}
where $A_t \in \mathbb{R}^{n \times n}$ and $B_t \in \mathbb{R}^{n \times m}$ and $u_t \in \mathbb{R}^m$ is an unknown external disturbance. We assume for simplicity that the initial state is $x_0 = 0$, though it is easy to extend our results to arbitrary $x_0$. At each timestep, we receive a noisy linear observation $y_t \in \mathbb{R}^p$ of the state: 
\begin{equation} \label{observation model}
y_t = C_t x_t + v_t,
\end{equation}
where $C_t \in \mathbb{R}^{p \times n}$ is a measurement matrix and  $v_t \in \mathbb{R}^p$ is an unknown measurement disturbance. The matrix $C_t$ is potentially sparse or low-rank, so that $y_t$ contains only limited information about the underlying state.
Our goal is to estimate
\begin{equation} \label{estimation-target}
    s_t = L_t x_t,
\end{equation}
where $L_t \in \mathbb{R}^{r \times n}$. We formulate estimation as an optimization problem where we seek to design an estimator $\hat{s}_t$ which minimizes the error 
\begin{equation} \label{estimation-error}
\sum_{t=0}^{T-1} \|\hat{s}_t - s_t \|_2^2.
\end{equation}
We assume that $\{A_t, B_t, C_t, L_t\}_{t=0}^{T-1}$ are known, so the only uncertainty in the evolution of the system comes from the disturbances $u, v$.

In the control setting, we consider a Linear Quadratic Regulator model where a system evolves according to the linear dynamics 
\begin{equation} \label{evolution-eq}
x_{t+1} = A_tx_t + B_{u, t} u_t + B_{w, t} w_t.
\end{equation}
Here $x_t \in \mathbb{R}^n$ is a state variable we seek to regulate, $u_t \in \mathbb{R}^m$ is a control variable which we can dynamically adjust to influence the evolution of the system, and $w_t \in \mathbb{R}^p$ is an external disturbance. We assume for simplicity that the initial state is $x_0 = 0$, though it is easy to extend our results to arbitrary $x_0$. We formulate the problem of controlling the system as an optimization problem, where the goal is to select the control actions so as to minimize the quadratic cost 
\begin{equation} \label{lqr-cost}
\sum_{t=0}^{T-1} \left( x_t^{\top}Q_t x_t + u_t^{\top}R_t u_t \right),
\end{equation}
where $Q_t \succeq 0, R_t \succ 0$ for $t = 0, \ldots T - 1$.  We assume that the dynamics $\{A_t, B_{u, t}, B_{w, t}\}_{t=0}^{T-1}$ and costs $\{ Q_t, R_t\}_{t=0}^{T-1}$ are known, so the only uncertainty in the evolution of the system comes from the disturbance $w$.

As is standard in the input-output approach to estimation and control, we encode estimators and controller as linear \textit{transfer operators} mapping the disturbances to the signal we wish to regulate. Let $u = (u_0, \ldots u_{T-1})$ and define $v, y, s, w, x$ analogously. We define the \textit{energy} in a signal $u = (u_0, \ldots u_{T-1})$ to be $$\|u\|_2^2 = \sum_{t = 0}^{T-1} \|u_t\|_2^2.$$ 

In the control setting, the dynamics (\ref{estimation-dynamics-model}) and observation model (\ref{observation model}) can be encoded as $$y_t  = \mathcal{H}u + v,$$ where $\mathcal{H}$ is an appropriately defined operator depending on $\{A_t, B_t, C_t\}_{t=0}^{T-1}$. Similarly, (\ref{estimation-target}) can be written as $$s = \mathcal{L}u,$$ where the operator $\mathcal{L}$ is block-diagonal with entries $\{L_t \}_{t=0}^{T-1}$, and the error (\ref{estimation-error}) is simply $$\|\hat{s} - s\|_2^2. $$

In the control setting, we wish to regulate the signal $s_t = Q_t^{1/2}x_t$, while simultaneously minimizing the energy in the control signal. The dynamics (\ref{evolution-eq}) are captured by the relation $$s = \mathcal{F}u + \mathcal{G}w, $$ where $\mathcal{F}$ and $\mathcal{G}$ are strictly causal operators encoding $\{A_t, B_{u, t}, B_{w, t}, Q_t^{1/2}\}_{t=0}^{T-1}$, and the LQR cost (\ref{lqr-cost}) can be written as $$ \|s\|_2^2 + \|u\|_2^2.$$ We note that this may involve re-parameterizing the original dynamics (\ref{evolution-eq}); we refer the reader to \cite{hassibi1999indefinite} for more background on transfer operators and the input-output approach to estimation and control.

\subsection{Robust estimation and control} \label{robust-estimation-control-sec}
Our results rely heavily on techniques from robust estimation and control; in particular, we show that the problem of finding a regret-optimal estimator can be reduced to an $H_{\infty}$ estimation problem. Similarly, the problem of finding a regret-optimal controller can be reduced to an $H_{\infty}$ control problem. In this section, we review the $H_{\infty}$ formulation of estimation and control problems, along with their state-space solutions.

\begin{problem}[$H_{\infty}$-optimal estimation]\label{hinf-optimal-estimation-problem}
Find a causal estimator that minimizes $$\sup_{w} \frac{\|\hat{s} - s \|_2^2}{\|u\|_2^2 + \|v\|_2^2}. $$
\end{problem}

This problem has the natural interpretation of minimizing the worst-case gain from the energy in the disturbances $u$ and $v$ to the error of the estimator. In general, it is not known how to derive a closed-form for the $H_{\infty}$-optimal estimator, so is it common to consider a relaxation: 

\begin{problem}[Suboptimal $H_{\infty}$ estimation at level $\gamma$] \label{hinf-suboptimal-estimation-problem} 
Given  $\gamma > 0$, find a causal estimator such that $$\|\hat{s} - s \|_2^2 < \gamma^2(\|u\|_2^2 + \|v\|_2^2)$$ for all disturbances $u, v$, or determine whether no such estimator exists.
This problem can also be expressed in terms of transfer operators; an equivalent formulation is to find $\hat{s} = \mathcal{K}y$ such that
\begin{align}
&(\mathcal{K}\Delta - \mathcal{L}\mathcal{H}^*\Delta^{-*})(\mathcal{K}\Delta - \mathcal{L}\mathcal{H}^*\Delta^{-*})^* \nonumber \\
& \hspace{5mm} + \mathcal{L}(I+\mathcal{H}^*\mathcal{H})^{-1}\mathcal{L}^* \prec \gamma^2I, \label{hinf-suboptimal-operator-condition}
\end{align} 
where $\Delta$ is the unique causal operator such that $$\Delta \Delta^* = I+\mathcal{H}\mathcal{H}^*. $$
\end{problem}
This problem has a well-known state-space solution:

\begin{theorem}[Theorem 4.2.1 and Lemma 4.2.9 in \cite{hassibi1999indefinite}] \label{hinf-suboptimal-estimation-thm}
Given $\gamma > 0$, a causal $H_{\infty}$ estimator at level $\gamma$ exists if and only if the matrices  $$\Sigma_t = \begin{bmatrix} I & 0 \\ 0 & -\gamma^2I \end{bmatrix}  + \begin{bmatrix} C_t \\ L_t \end{bmatrix}P_t \begin{bmatrix} C_t^{\top} & L_t^{\top} \end{bmatrix} $$ and $$\begin{bmatrix} I & 0 \\ 0 & -\gamma^2I \end{bmatrix}$$ have the same inertia, where $P_t$ is the solution of the Riccati recursion $$P_{t+1} = A_tP_tA_t^{\top} + B_tB_t^{\top} - A_tP_t \begin{bmatrix} C_t^{\top} & L_t^{\top} \end{bmatrix} \Sigma_t^{-1}\begin{bmatrix} C_t \\ L_t \end{bmatrix}.$$
In this case, one possible $H_{\infty}$ estimator is given by $$\hat{s}_t = L_t \hat{x}_{t \mid t}, $$ where $x_{t \mid t}$ is recursively computed as $$\hat{x}_{t+1 \mid t+1} = A_t \hat{x}_{t \mid t} + K_{t+1}(y_{t+1} - C_{t+1}\hat{x}_{t \mid} t), $$ where we initialize $\hat{x}_0 = x_0$ and define $$K_t = P_tC_t^{\top}(I + C_tP_tC_t^{\top})^{-1}.$$ 
\end{theorem}

We emphasize that an $H_{\infty}$-optimal estimator is easily obtained from a solution of the suboptimal $H_{\infty}$ estimation problem by bisection on $\gamma$.

\begin{problem}[$H_{\infty}$-optimal control] \label{hinf-optimal-control-problem}
Find a causal controller that minimizes $$\sup_{w} \frac{\|s\|_2^2 + \|u\|_2^2}{\|w\|_2^2}. $$
\end{problem}
This problem has the natural interpretation of minimizing the worst-case gain from the energy in the disturbance $w$ to the cost incurred by the controller. In general, it is not known how to derive a closed-form for the $H_{\infty}$-optimal controller, so instead is it common to consider a relaxation: 

\begin{problem} [Suboptimal $H_{\infty}$ control at level $\gamma$] \label{hinf-suboptimal-control-problem}
Given  $\gamma > 0$, find a causal controller such that  $$\|s\|_2^2 + \|u\|_2^2 < \gamma^2 \| w\|_2^2 $$ for all disturbances $w$, or determine whether no such controller exists.
\end{problem}

This problem has a well-known state-space solution:
\begin{theorem}[Theorem 9.5.1 in \cite{hassibi1999indefinite}] \label{hinf-suboptimal-controller-thm}
Given $\gamma > 0$, an $H_{\infty}$ controller at level $\gamma$ exists if and only if
 $$ B_{w, t}^{\top}P_{t+1}B_{w, t} - B_{w, t}^{\top}P_{t+1}B_{u, t} H_t^{-1}B_{u, t}^{\top}P_{t+1}B_{w, t} \prec -\gamma^2I $$
for all $t = 0, \ldots T-1$, where
$$H_t = R_t + B_{u, t}^{\top}P_{t+1}B_{u, t},$$
we define $P_t$ to be the solution of the backwards-time Riccati equation $$P_t = Q_t + A_t^{\top}P_{t+1}A_t -  A_t^{\top}P_{t+1}\hat{B}_t \hat{H}_t^{-1} \hat{B}_t^{\top} P_{t+1}A_t $$ with initialization $P_{T-1} = Q_{T-1}$, and we define $$ \hat{B}_t = \begin{bmatrix} B_{u, t} & B_{w, t} \end{bmatrix}, $$ $$\hat{H}_t = \begin{bmatrix} R_t & 0 \\ 0 & -\gamma^2 I \end{bmatrix}  + \hat{B}_t^{\top} P_{t+1} \hat{B}_t.$$
In this case, the suboptimal $H_{\infty}$ controller has the form $$ u_t =  -H_t^{-1} B_{u, t}^{\top}P_{t+1}(A_tx_t +  B_{w, t}  w_t).$$
\end{theorem}
\vspace{2mm}
As in the estimation setting, an $H_{\infty}$-optimal controller is easily obtained from the solution of the suboptimal $H_{\infty}$ control problem by bisection on $\gamma$.

\subsection{The noncausal estimator and the noncausal controller} \label{noncausal-sec}
We briefly state a few facts about the clairvoyant noncausal estimator and the clairvoyant noncausal controller (sometimes called the offline optimal controller). 
The optimal noncausal estimate of $s = \mathcal{L}u$ given the observation $y = \mathcal{H}u + v$ is  $$\hat{s}_{nc} = \mathcal{L}\mathcal{H}^*(I + \mathcal{H}\mathcal{H}^*)^{-1}y,$$ and the corresponding error  $\|\hat{s}_{nc} - s\|_2^2$ is
\begin{equation} \label{noncausal-estimation-cost}
 \left\| \mathcal{L}u - \mathcal{L}\mathcal{H}^*(I + \mathcal{H}\mathcal{H}^*)^{-1}(\mathcal{H}u + v)\right\|_2^2.
\end{equation}
The optimal noncausal control action in response to the input disturbance $w$ is given by
\begin{equation*}  \label{offline-operator}
u = - (I + \mathcal{F}^*\mathcal{F})^{-1}\mathcal{F}^*\mathcal{G}w.
\end{equation*}
and the corresponding cost incurred by the noncausal controller is
\begin{equation} \label{offline-operator-cost}
 w^{\top}\mathcal{G}^*(I + \mathcal{F}\mathcal{F}^*)^{-1}\mathcal{G}w.
\end{equation}
We refer to Theorem 10.3.1 and Theorem 11.2.1 in \cite{hassibi1999indefinite} for details. For a state-space description of the noncausal controller, see \cite{goel2020power}.

\subsection{Regret-optimal estimation and control}
In this paper, instead of minimizing the worst-case \textit{cost} as in traditional $H_{\infty}$ estimation and control, our goal is to design estimators and controllers that minimize the worst-case \textit{regret}.  
The regret of a causal estimator $\hat{s}$ on inputs $u, v$ is the difference in the cost it incurs and the cost that a clairvoyant noncausal estimator $\hat{s}_{nc}$ would incur. In light of (\ref{noncausal-estimation-cost}), the regret is 
\begin{equation} \label{regret-estimation}
\|\hat{s} - s\|_2^2 - \left\| \mathcal{L}u - \mathcal{L}\mathcal{H}^*(I + \mathcal{H}\mathcal{H}^*)^{-1}(\mathcal{H}u + v)\right\|_2^2.
\end{equation}
Similarly, the regret of a causal controller on an input disturbance $w$ is the difference in the cost it incurs and the cost that a clairvoyant noncausal controller would incur:
\begin{equation} \label{regret-control}
 \|s\|_2^2 + \|u\|_2^2 - w^{\top}\mathcal{G}^*(I + \mathcal{F}\mathcal{F}^*)^{-1}\mathcal{G}w,
\end{equation}
where $s = \mathcal{F}u + \mathcal{G}w $ and $u$ is the sequence of control actions selected by the causal controller.
It is natural to formulate regret-minimization analogously to the $H_{\infty}$ problems considered in Section \ref{robust-estimation-control-sec}:

\begin{problem}[Regret-optimal estimation]\label{regret-optimal-estimation-problem}
Find a causal estimator that minimizes $$\sup_{w} \frac{\|\hat{s} - s \|_2^2 - \left\| \mathcal{L}u - \mathcal{L}\mathcal{H}^*(I + \mathcal{H}\mathcal{H}^*)^{-1}(\mathcal{H}u + v)\right\|_2^2.}{\|u\|_2^2 + \|v\|_2^2}. $$ 
\end{problem}

\vspace{3mm}
\noindent This problem has the natural interpretation of minimizing the worst-case gain from the energy in the disturbances $u, v$ to the regret incurred by the estimator.
\begin{problem}[Regret-suboptimal estimation at level $\gamma$]\label{regret-suboptimal-estimation-problem} 
Given  $\gamma > 0$, find a causal estimator such that 
\begin{align*}
& \|\hat{s} - s \|_2^2 - \left\| \mathcal{L}u - \mathcal{L}\mathcal{H}^*(I + \mathcal{H}\mathcal{H}^*)^{-1}(\mathcal{H}u + v)\right\|_2^2 \\
& \hspace{5mm} < \gamma^2(\|u\|_2^2 + \|v\|_2^2)    
\end{align*}
for all disturbances $u, v$, or determine whether no such estimator exists.
\end{problem}

\begin{problem}[Regret-optimal control at level $\gamma$] \label{regret-optimal-control-problem}
Find a causal controller that minimizes $$\sup_{w} \frac{\|s\|_2^2 + \|u\|_2^2 - w^{\top}\mathcal{G}^*(I + \mathcal{F}\mathcal{F}^*)^{-1}\mathcal{G}w}{\|w\|_2^2}. $$
\end{problem}
\vspace{3mm}
\noindent This problem has the natural interpretation of minimizing the worst-case gain from the energy in the disturbance $w$ to the regret incurred by the controller. As in the $H_{\infty}$ setting, we consider the relaxation:

\begin{problem} [Regret-suboptimal control] \label{regret-suboptimal-control-problem}
Given a performance level $\gamma > 0$, find a causal controller such that $$ \|s\|_2^2 + \|u\|_2^2 - w^{\top}\mathcal{G}^*(I + \mathcal{F}\mathcal{F}^*)^{-1}\mathcal{G}w < \gamma^2\|w\|_2^2  $$ for all disturbances $w$, or determine whether no such policy exists.
\end{problem}

We emphasize that, as in the $H_{\infty}$ setting, if we can solve the regret-suboptimal estimation and control problems, we can easily recover solutions to the regret-optimal estimation and control problems by bisection.

% Equation (\ref{offline-operator}) gives the optimal noncausal controller in input-output form; recently \cite{goel2020power} and \cite{foster2020logarithmic} obtained a state-space description using dynamic programming:

% \begin{theorem}[\cite{goel2020power}] \label{offline-structure-thm}
% The optimal noncausal controller has the form $$u_t^* = -H_t^{-1}B_{u, t}^{\top} \left(P_{t+1}A_t x_t + P_{t+1}B_{w, t} w_t + \frac{1}{2}v_{t+1} \right),$$ where we define $$H_t = R_t + B_{u, t}^{\top}P_{t+1}B_{u, t}, $$ $P_t$ is the solution of the Riccati recurrence (\ref{Riccati-recur}), and $v_t$ satisfies the recurrence 
% $$ v_t = 2A_t^{\top}S_t B_{w, t} w_t + A_t^{\top}S_tP_{t+1}^{-1}v_{t+1}, $$
% and we define 
% $$S_t = P_{t+1} - P_{t+1}B_{u, t}H_t^{-1}B_{u, t}^{\top}P_{t+1}.$$
% \end{theorem}
% In light of Theorem \ref{h2-optimal-control-theorem}, this shows that \textit{the regret-optimal control action at time $t$ is the sum of the $H_2$-optimal control action and a linear combination of the future disturbances $w_{t+1}, \ldots w_{T-1}$.}

\section{Regret-optimal estimation}
We seek to design a causal estimator (a filter) $\hat{s} = \mathcal{K}y$ such that the error residual $\tilde{s} = s - \hat{s}$ has the smallest possible energy relative to the error residual associated with the noncausal estimator. Our approach is to reduce the regret-suboptimal estimation problem (Problem \ref{regret-suboptimal-estimation-problem}) to the suboptimal $H_{\infty}$ estimation problem (Problem \ref{hinf-suboptimal-estimation-problem}).

For any operator $\mathcal{K}$, the corresponding transfer operator $$\mathcal{T}_{\mathcal{K}} : \begin{bmatrix} u \\ v \end{bmatrix} \rightarrow \tilde{s} $$ is given by $\begin{bmatrix} \mathcal{L}- \mathcal{K}\mathcal{H} & -\mathcal{K}  \end{bmatrix}. $
Recall that the optimal noncausal estimator is $$\mathcal{K}_{nc} = \mathcal{L}\mathcal{H}^*(I + \mathcal{H}\mathcal{H}^*)^{-1}, $$
 and the regret-suboptimal estimation problem at level $\gamma$ is to find a causal estimator $\hat{s}$ such that 
\begin{equation} \label{regret-cond}
\|\hat{s} - s\|^2 - \|\hat{s}_{nc} - s\|^2 < \gamma^2 (\|u\|^2 + \|v\|^2),
\end{equation}
or to determine whether no such estimator exists. Condition (\ref{regret-cond}) can be neatly expressed in terms of transfer operators as 
\begin{equation} \label{regret-cond-transfer-operator}
0 \prec \gamma^2 I + \mathcal{T}_{\mathcal{K}_{nc}}^* \mathcal{T}_{\mathcal{K}_{nc}} - \mathcal{T}_{\mathcal{K}}^* \mathcal{T}_{\mathcal{K}}.
\end{equation}
Define the unitary operator $$\theta = \begin{bmatrix} I & \mathcal{H}^* \\ -\mathcal{H} & I \end{bmatrix} \begin{bmatrix} \Delta_1^{-1} & 0 \\ 0 & -\Delta_2^{-*} \end{bmatrix},$$
where we define causal operators $\Delta_1, \Delta_2$ such that $$ \Delta_1^*\Delta_1 = I + \mathcal{H}^*\mathcal{H}, \hspace{5mm} \Delta_2 \Delta_2^* = I + \mathcal{H}\mathcal{H}^*. $$
Notice that for every estimator $\mathcal{K}$, we have $$\mathcal{T}_{\mathcal{K}}\theta =  \begin{bmatrix} \mathcal{L}\Delta_1^{-1} & \mathcal{K}\Delta_2 - \mathcal{L}\mathcal{H}^*\Delta_2^{-*}  \end{bmatrix},$$ and in particular, we have $$\mathcal{T}_{\mathcal{K}_{nc}}\theta =  \begin{bmatrix} \mathcal{L}\Delta_1^{-1} & 0 \end{bmatrix}.$$
Since $\theta$ is unitary, condition (\ref{regret-cond-transfer-operator}) is equivalent to 
\begin{equation} 
0 \prec \gamma^2 I + \theta^*\mathcal{T}_{\mathcal{K}_{nc}}^* \mathcal{T}_{\mathcal{K}_{nc}}\theta - \theta^*\mathcal{T}_{\mathcal{K}}^* \mathcal{T}_{\mathcal{K}}\theta.
\end{equation}
Expanding the right-hand side and using the the Schur complement, we see that this condition is equivalent to 
\begin{equation*} 
(\mathcal{S}\mathcal{K}\Delta_2 - \mathcal{S}\mathcal{L}\mathcal{H}^*\Delta_2^{-*})^* (\mathcal{S}\mathcal{K}\Delta_2 - \mathcal{S}\mathcal{L}\mathcal{H}^*\Delta_2^{-*}) \prec \gamma^2 I,
\end{equation*}
where $\mathcal{S}$ is the unique causal operator such that $$\mathcal{S}^{*}\mathcal{S} = I + \gamma^{-2}\mathcal{L}(I + \mathcal{H}^*\mathcal{H})^{-1}\mathcal{L}^*.$$ 
This condition is equivalent to 
\begin{equation*} 
(\mathcal{S}\mathcal{K}\Delta_2 - \mathcal{S}\mathcal{L}\mathcal{H}^*\Delta_2^{-*}) (\mathcal{S}\mathcal{K}\Delta_2 - \mathcal{S}\mathcal{L}\mathcal{H}^*\Delta_2^{-*})^* \prec \gamma^2 I.
\end{equation*} 
Left-multiplying by $\mathcal{S}^{-1}$ and right-multiplying by $\mathcal{S}^{-*}$, we obtain
\begin{equation*} 
(\mathcal{K}\Delta_2 - \mathcal{L}\mathcal{H}^*\Delta_2^{-*}) (\mathcal{K}\Delta_2 - \mathcal{L}\mathcal{H}^*\Delta_2^{-*})^* \prec \gamma^2 (\mathcal{S}^*\mathcal{S})^{-1}.
\end{equation*} 
Adding $\mathcal{L}(I + \mathcal{H}^*\mathcal{H})^{-1}\mathcal{L}^*$ to both sides we obtain
\begin{align} 
& (\mathcal{K}\Delta_2 - \mathcal{L}\mathcal{H}^*\Delta_2^{-*}) (\mathcal{K}\Delta_2 - \mathcal{L}\mathcal{H}^*\Delta_2^{-*})^* + \mathcal{L}(I + \mathcal{H}^*\mathcal{H})^{-1}\mathcal{L}^* \nonumber \\
& \hspace{5mm} \prec \gamma^2 (\mathcal{S}^*\mathcal{S})^{-1} + \mathcal{L}(I + \mathcal{H}^*\mathcal{H})^{-1}\mathcal{L}^*. \label{before-t-cond}
\end{align} 
Let $\mathcal{T}$ be the unique causal operator such that 
\begin{equation} \label{t-factorization}
\mathcal{T} \mathcal{T}^* = \gamma^2 (\mathcal{S}^*\mathcal{S})^{-1} + \mathcal{L}(I + \mathcal{H}^*\mathcal{H})^{-1}\mathcal{L}^*.
\end{equation}
Left-multiplying (\ref{before-t-cond}) by $\mathcal{T}^{-1}$ and right-multiplying by $\mathcal{T}^{-*}$, we obtain
\begin{align}
& (\mathcal{T}^{-1}\mathcal{K}\Delta_2 - \mathcal{T}^{-1}\mathcal{L}\mathcal{H}^*\Delta_2^{-*}) (\mathcal{T}^{-1}\mathcal{K}\Delta_2 - \mathcal{T}^{-1}\mathcal{L}\mathcal{H}^*\Delta_2^{-*})^* \nonumber \\
& \hspace{5mm}  + \mathcal{T}^{-1}\mathcal{L}(I + \mathcal{H}^*\mathcal{H})^{-1}\mathcal{L}^*\mathcal{T}^{-*} \prec I.  \label{final-condition}
\end{align} 
Comparing with Problem (\ref{hinf-suboptimal-estimation-problem}), we see that condition (\ref{final-condition}) is in the form (\ref{hinf-suboptimal-operator-condition}) with $\mathcal{L}' = \mathcal{T}^{-1}\mathcal{L}$, $\mathcal{K}' = \mathcal{T}^{-1}\mathcal{K}$, $\Delta' = \Delta_2$, and $\gamma = 1$. The main technical challenge to deriving the regret-suboptimal estimator in state-space form is to obtain the factorization (\ref{t-factorization}); with this in hand, we obtain the following theorem.
\begin{theorem} \label{regret-optimal-estimation-thm}
The regret-optimal filter is given by
$$ \hat{s}_t = L_t \hat{x}_{t \mid t},$$ where the state estimates can be recursively computed as $$\begin{bmatrix} \hat{x}_{t+1 \mid t+1} \\ \hat{\eta}_{t + 1\mid t + 1} \end{bmatrix} = \hat{A}_t \begin{bmatrix} \hat{x}_{t\mid t} \\ \hat{\eta}_{t \mid t} \end{bmatrix} + \hat{K}_{t+1} (y_{t+1} - C_{t+1}A_t \hat{x}_{t\mid t} ),$$  and we initialize $\hat{x}_{0 \mid 0} = 0, \hat{\eta}_{0 \mid 0} = 0$. We define 
$$ \hat{A}_t = \begin{bmatrix} A_t & 0  \\ \tilde{K}_t L_t & \tilde{A}_t - \tilde{K}_t \tilde{C}_t \end{bmatrix},$$ $$ \hat{B}_t = \begin{bmatrix}  B_t \\ 0 \end{bmatrix},  \hspace{5mm} \hat{L}_t = \begin{bmatrix} \tilde{\Sigma}_t^{-1/2}L_t & -\tilde{\Sigma}_t^{-1/2}\tilde{C}_t \end{bmatrix}, $$ $$\hat{\Sigma}_t = \begin{bmatrix} I & 0 \\ 0 & -I \end{bmatrix} + \begin{bmatrix} \hat{C}_t \\ \hat{L}_t \end{bmatrix}  \hat{P}_t \begin{bmatrix} \hat{C}_t^{\top} & \hat{L}_t^{\top} \end{bmatrix},$$
$$\hat{K}_t = \hat{P}_t\hat{C}_t^{\top}(I + \hat{C}_t\hat{P}_t\hat{C}_t^{\top})^{-1},$$
and $\hat{P}_t$ is the solution of the Riccati recursion $$\hat{P}_{t+1} = \hat{A}_t \hat{P}_t \hat{A}_t^{\top} + \hat{B}_t\hat{B}_t^{\top} -  \hat{A}_t \hat{P}_t \begin{bmatrix} \hat{C}_t^{\top} & \hat{L}_t^{\top} \end{bmatrix}\hat{\Sigma}^{-1}_t \begin{bmatrix} \hat{C}_t \\ \hat{L}_t \end{bmatrix} \hat{P}_t  \hat{A}_t^{\top},$$ where we initialize $\hat{P}_{T-1} = 0$. The matrices $\tilde{A}_t, \tilde{B}_t, \tilde{C}_t, \tilde{L}_t, \tilde{K}_t,$ and $\tilde{\Sigma}_t$ are defined in (\ref{tilde-matrices-estimation}) and (\ref{tilde-matrices-estimation2}). The regret-optimal filter is the regret-suboptimal filter at level $\gamma_{opt}$, where $\gamma_{opt}$ is the smallest value of $\gamma$ such that $\hat{\Sigma}_t$ has the same inertia as the matrix $ \begin{bmatrix} I & 0 \\ 0 & -I \end{bmatrix}$. The regret incurred by the regret-optimal filter is at most $\gamma_{opt}^2 (\|u\|_2^2 + \|w\|_2^2)$.
\end{theorem}
\proof{See the Appendix.}

\section{Regret-optimal Control} \label{regret-optimal-control-sec}
We now turn to the problem of deriving the regret-optimal controller. Our approach is to reduce the  regret-suboptimal control problem (Problem \ref{regret-suboptimal-control-problem})  to the suboptimal $H_{\infty}$ problem (Problem \ref{hinf-suboptimal-control-problem}); once the regret-suboptimal controller is found, the regret-optimal controller can be easily obtained by bisection on $\gamma$. Recall that the regret-suboptimal problem at level $\gamma$ is to find, if possible, a causal control policy such that for all disturbances $w$, 
$$ \|s\|_2^2 + \|u\|_2^2 - w^{\top}\mathcal{G}^*(I + \mathcal{F}\mathcal{F}^*)^{-1}\mathcal{G}w < \gamma^2\|w\|_2^2,  $$
or equivalently
\begin{equation} \label{regret-suboptimal-problem-2}
\|u\|_2^2 + \|s\|_2^2 < w^{\top} (\gamma^2 I +  \mathcal{G}^*(I + \mathcal{F}\mathcal{F}^*)^{-1}\mathcal{G})w,
\end{equation}
where $s = \mathcal{F}u + \mathcal{G}w$.
Since $\gamma^2 I +  \mathcal{G}^*(I + \mathcal{F}\mathcal{F}^*)^{-1}\mathcal{G}$ is strictly positive definite for all $\gamma > 0$, there exists a unique causal, invertible matrix $\Delta$ such that $$\gamma^2 I +  \mathcal{G}^*(I + \mathcal{F}\mathcal{F}^*)^{-1}\mathcal{G} = \Delta^*\Delta.$$ Notice that $$w^{\top} (\gamma^2 I +  \mathcal{G}^*(I + \mathcal{F}\mathcal{F}^*)^{-1}\mathcal{G})w = \|\Delta w\|_2^2.$$ Letting $w' = \Delta w$ and $\mathcal{G}' = \mathcal{G}\Delta^{-1}$, we have $s = \mathcal{F}u + \mathcal{G}' w'$. With this change of variables, the regret-suboptimal problem (\ref{regret-suboptimal-problem-2}) takes the form of finding, if possible, a causal controller such that for all $w'$, $$\|u\|_2^2 + \|s\|_2^2 < \|w'\|_2^2.$$
Comparing with Problem \ref{hinf-suboptimal-control-problem}, we see that this is a suboptimal $H_{\infty}$ problem at level $\gamma = 1$ in the system $s = \mathcal{F}u + \mathcal{G}' w'$. 

The main technical challenge to deriving a state-space model for the regret-suboptimal controller is to obtain an explicit factorization of $\gamma^2I + \mathcal{G}^*(I + \mathcal{F} \mathcal{F}^*)^{-1}\mathcal{G}$ as $\Delta^*\Delta$; once we have obtained $\Delta$ it is straightforward to recover a state-space description of the regret-optimal controller using the state-space description of the $H_{\infty}$ controller (Theorem \ref{hinf-suboptimal-controller-thm}). We obtain the following theorem:

\begin{theorem} \label{regret-optimal-controller-thm}
The regret-suboptimal controller at level $\gamma$ is given by
$$ u_t = - \hat{H}_t^{-1} \hat{B}_{u, t}^{\top} \hat{P}_{t+1} \begin{bmatrix} A_t \zeta_t + B_{w, t} w_t \\ \tilde{A}_t \nu_t + B_{w, t} w_t \end{bmatrix},$$
 where we define 
$$\hat{A}_t = \begin{bmatrix} A_t & - B_{w, t}(K^b_t)^{\top} \\ 0 &  \tilde{A}_t - B_{w, t}(K^b_t)^{\top} \end{bmatrix}, \hspace{5mm} \hat{B}_{u, t} =  \begin{bmatrix} B_{u, t} \\ 0 \end{bmatrix}, $$ $$ \hat{Q}_t = \begin{bmatrix} Q_t & 0 \\ 0 & 0 \end{bmatrix}, \hspace{5mm} \hat{H}_t = I + \hat{B}_{u, t}^{\top} \hat{P}_{t+1} \hat{B}_{u, t},$$ $$ \tilde{A}_t = A_t - K_t Q_t^{1/2} , \hspace{3mm} K_t = A_t P_t Q_t^{1/2} \Sigma_t^{-1}, $$ $$ \Sigma_t = I + Q_t^{1/2}P_t Q_t^{1/2}, \hspace{3mm} K_t^b = \tilde{A}_t^{\top}P_t^b B_{w, t} (\Sigma_t^b)^{-1}, $$ $$ \Sigma_t^b = \gamma^2 I + B^{\top}_{w, t} P_t^b B_{w, t},$$
 and the state variables $\zeta_t$ and $\nu_t$ evolve according to to the dynamics
$$\zeta_{t+1} = A_t \zeta_t + B_{u, t}u_t + B_{w, t}w_t, \hspace{5mm} \nu_{t+1} = \tilde{A}_t \nu_t + B_{w, t} w_t, $$
where we initialize $\zeta_0, \nu_0 = 0$. We define 
 $P_t$ to be the solution of the forwards Riccati recursion
 $$P_{t+1} = A_t P_t A_t^{\top} + B_{u, t}B_{u, t}^{\top} - K_t \Sigma_t K_t^{\top}, $$ where we initialize $P_0 = 0$, and define $\hat{P}_t, P_t^b$ to be the solutions of the backwards Riccati recursions 
$$ P_{t-1}^b =  \tilde{A}_t^{\top} P_{t}^b \tilde{A}_t + Q_t^{1/2}(\Sigma_t)^{-1}Q_t^{1/2} - K_t^b \Sigma_t^b (K_t^b)^{\top}, $$
$$\hat{P}_{t} = \hat{Q}_t + \hat{A}_t^{\top}\hat{P}_{t+1}A_t - \hat{A}_t^{\top} \hat{P}_{t+1} \hat{B}_{u, t} \hat{H}_t^{-1} \hat{B}_{u, t}^{\top} \hat{P}_{t+1} \hat{A}_t, $$ where we initialize $P^b_{T-1} = 0, \hat{P}_T = 0$. 

The regret-optimal causal controller is the regret-suboptimal controller at level $\gamma_{opt}$, where $\gamma_{opt}$ is the smallest value of $\gamma$ such that $$ \hat{B}_{w, t}^{\top} \hat{P}_{t+1} \hat{B}_{w, t} - \hat{B}_{w, t}^{\top} \hat{P}_{t+1}\hat{B}_{u, t} \hat{H}_t^{-1}\hat{B}_{u, t}^{\top}\hat{P}_{t+1} \hat{B}_{w, t} \prec \gamma^2 I$$ for all $t = 0, \ldots T-1$.
Furthermore, the regret incurred by the regret-optimal controller is at most $\gamma_{opt}^2 \|w\|_2^2$.
 \end{theorem}
\proof{ See the Appendix.}
We note that our results easily extend to settings where the controller has access to predictions of the next $k$ disturbances, or only affects the system dynamics after a delay of $k$ timesteps; we refer the reader to Section \ref{predictions-delay-sec} for details.

\section{Numerical Experiments} \label{numerical-experiments-sec}

\begin{figure}[h]
\centering
\includegraphics[width=0.9\columnwidth]{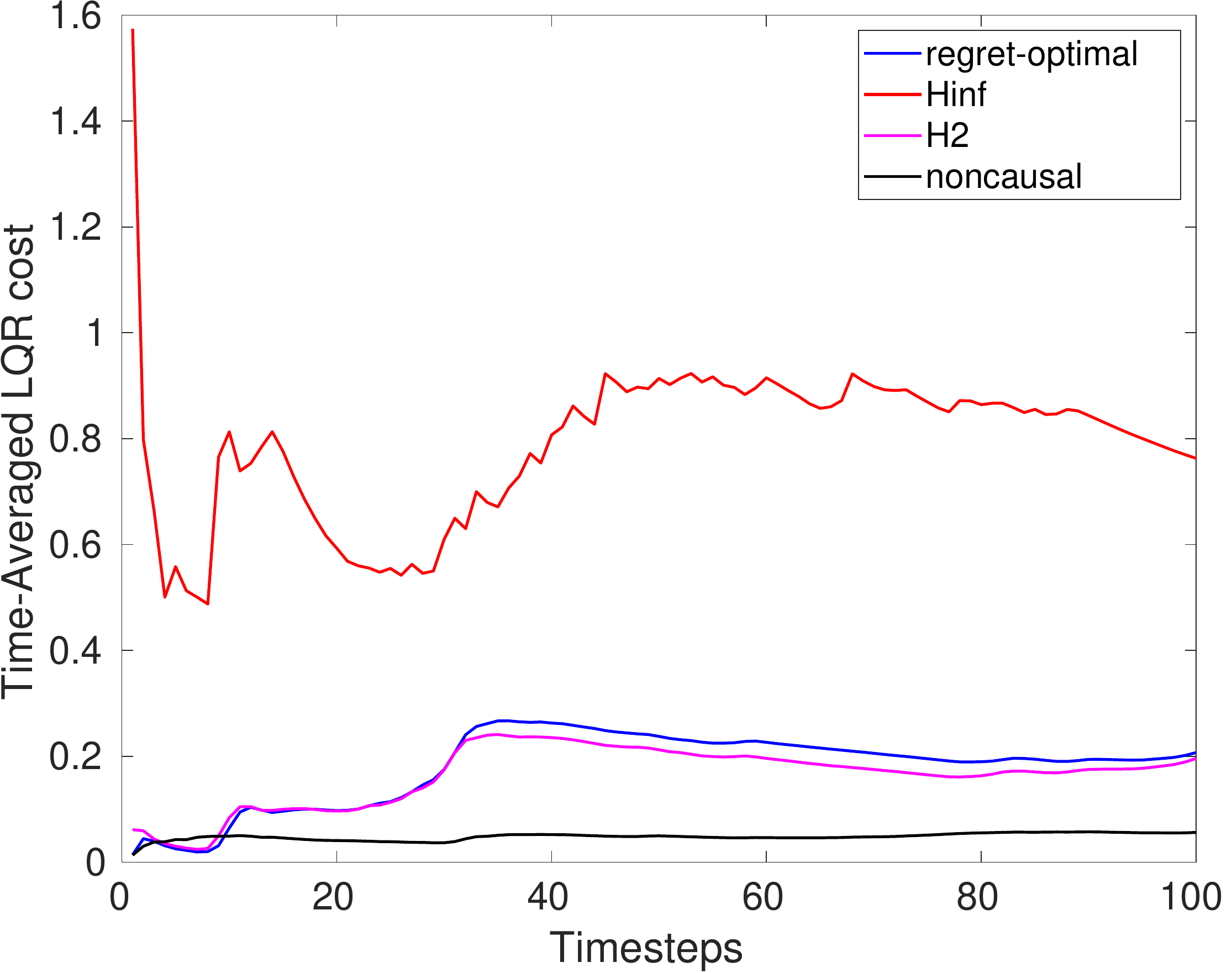}
\caption{Relative performance of LQR controllers with Gaussian noise.}
\label{gaussian-fig}
\end{figure}

\begin{figure}[h]
\centering
\includegraphics[width=0.9\columnwidth]{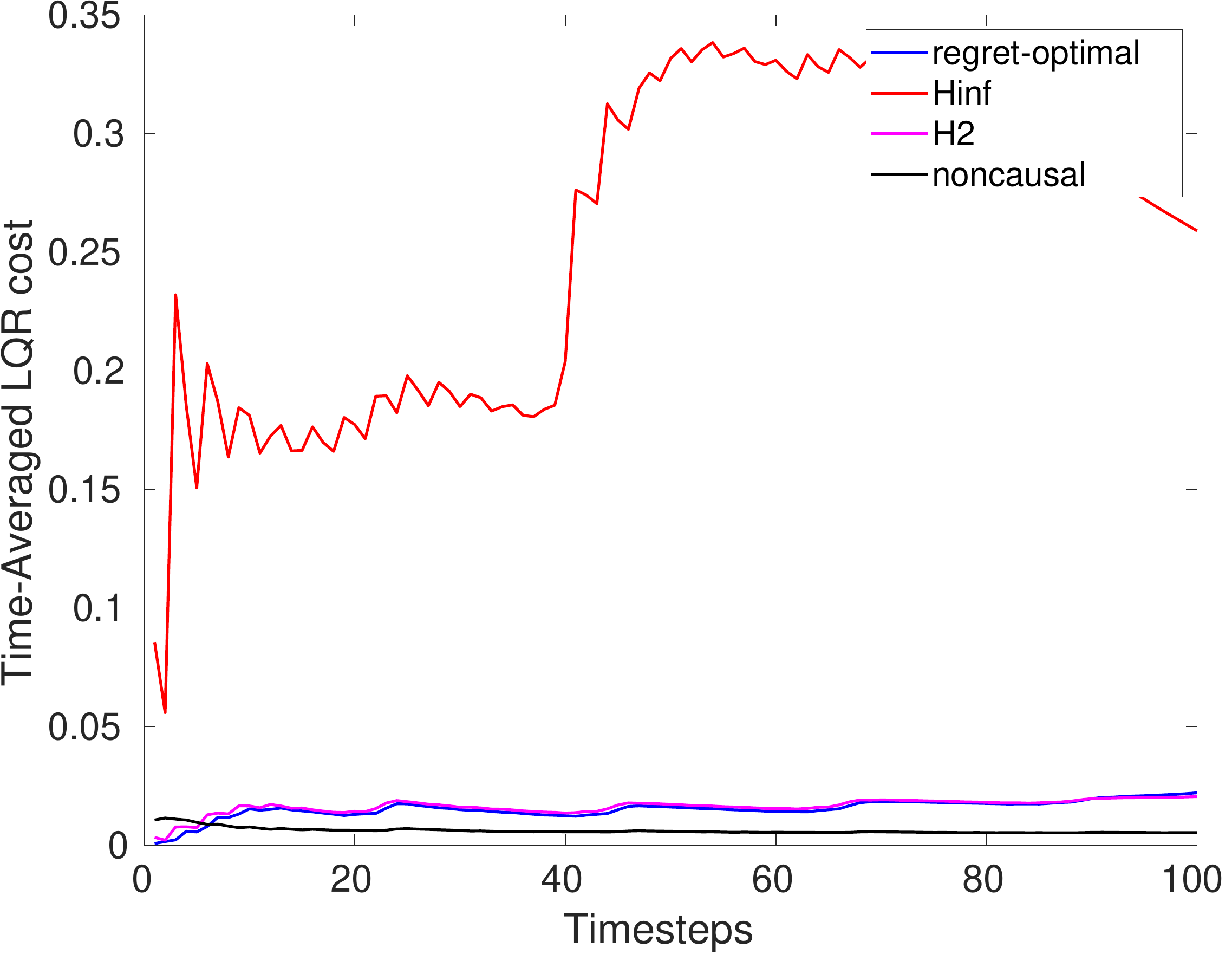}
\includegraphics[width=0.9\columnwidth]{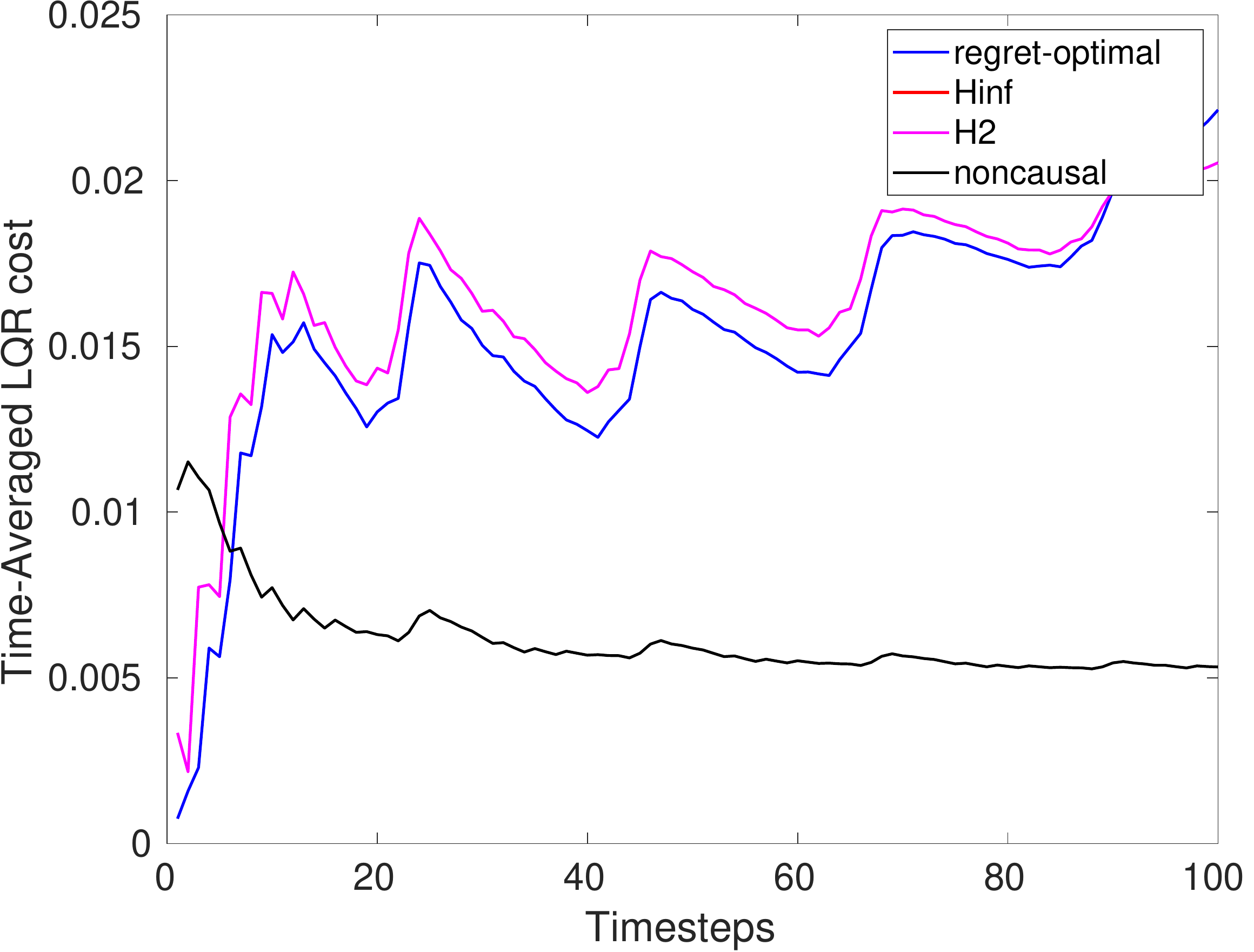}
\caption{Relative performance of LQR controllers with sawtooth noise.}
\label{sawtooth-fig}
\end{figure}

\begin{figure}[h]
\centering
\includegraphics[width=0.9\columnwidth]{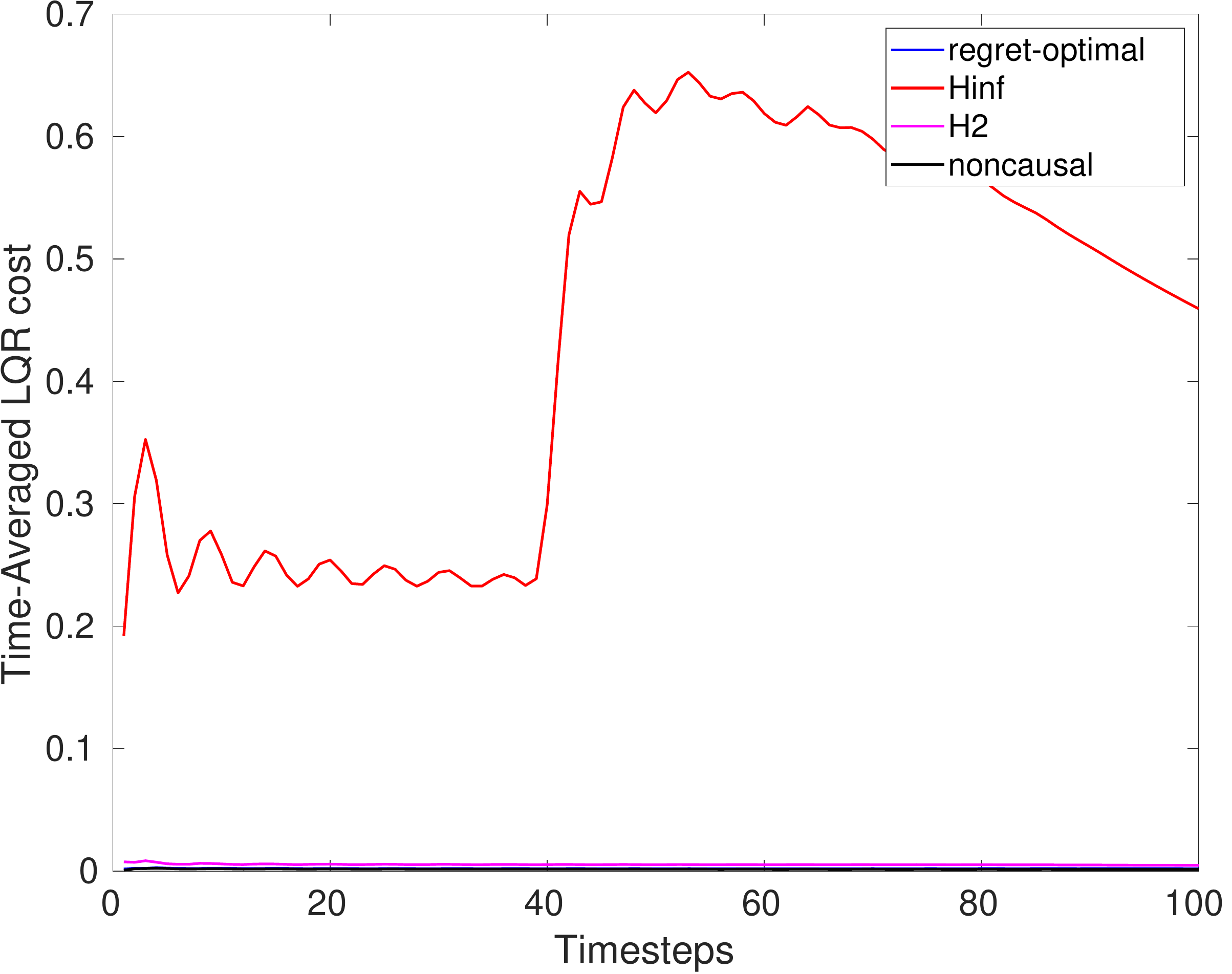}
\includegraphics[width=0.9\columnwidth]{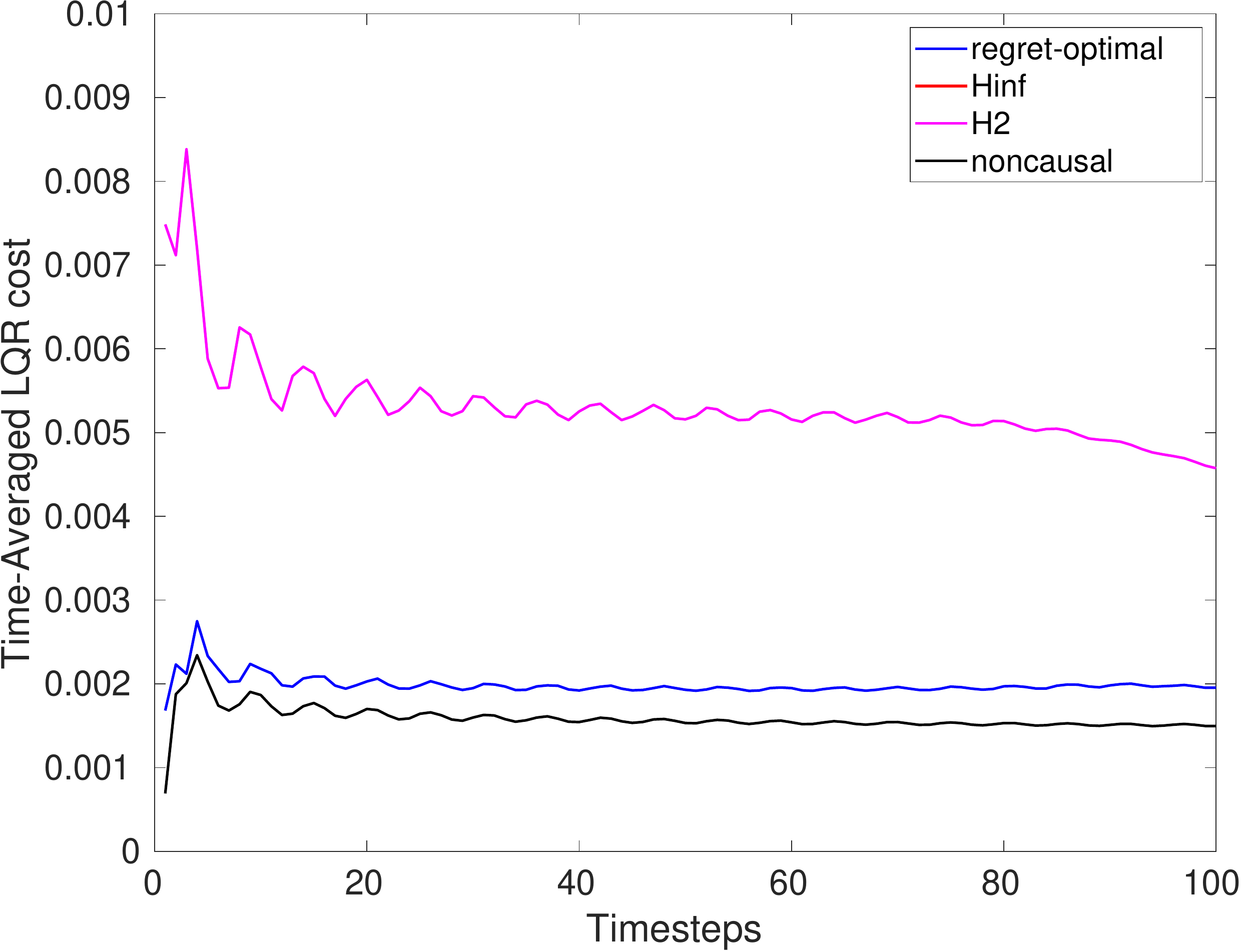}
\caption{Relative performance of LQR controllers with sinusoidal noise $w(k\Delta) = \sin{(10k\Delta)}$.}
\label{sinusoidal-fig-2}
\end{figure}

\begin{figure}[h]
\centering
\includegraphics[width=0.9\columnwidth]{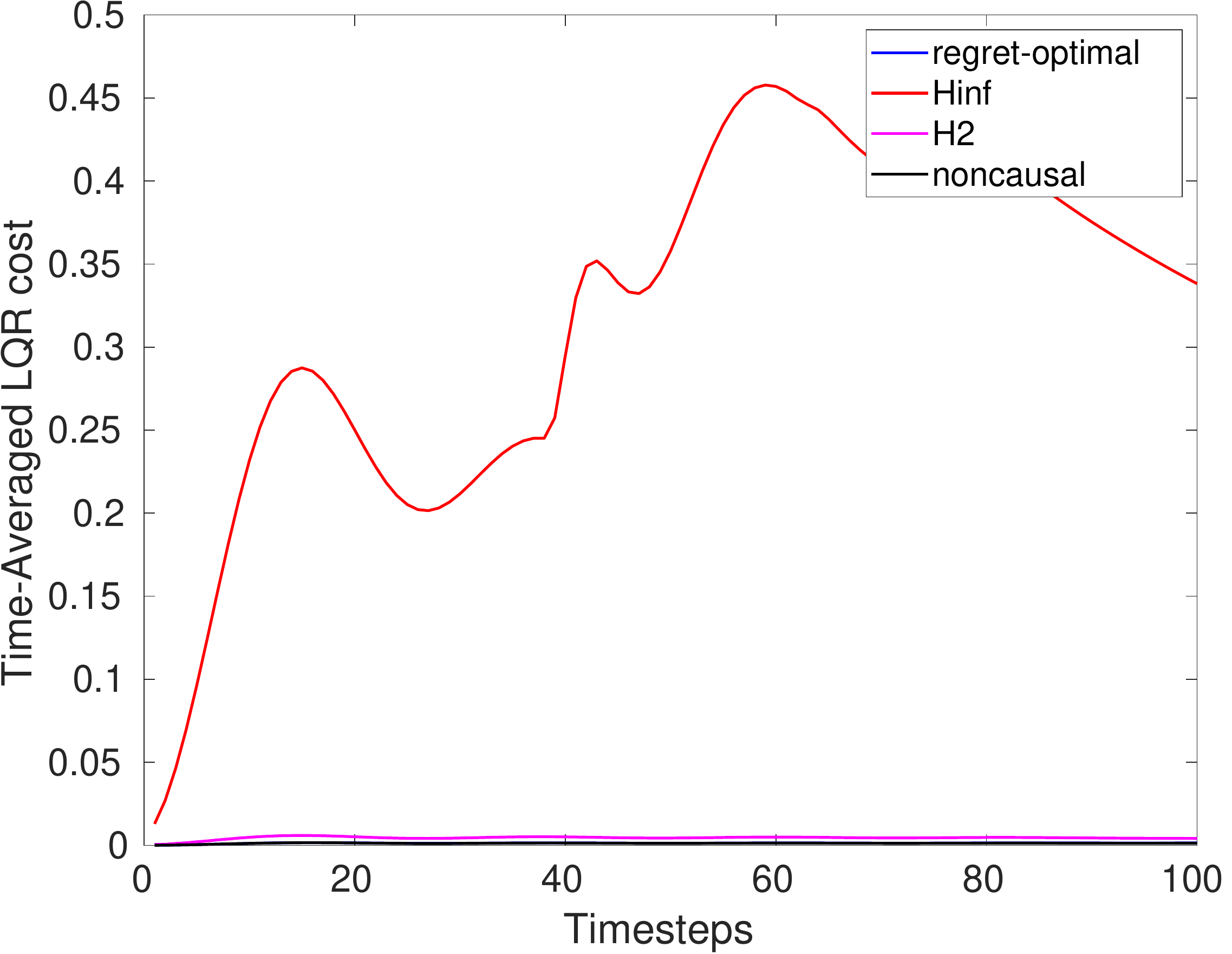}
\includegraphics[width=0.9\columnwidth]{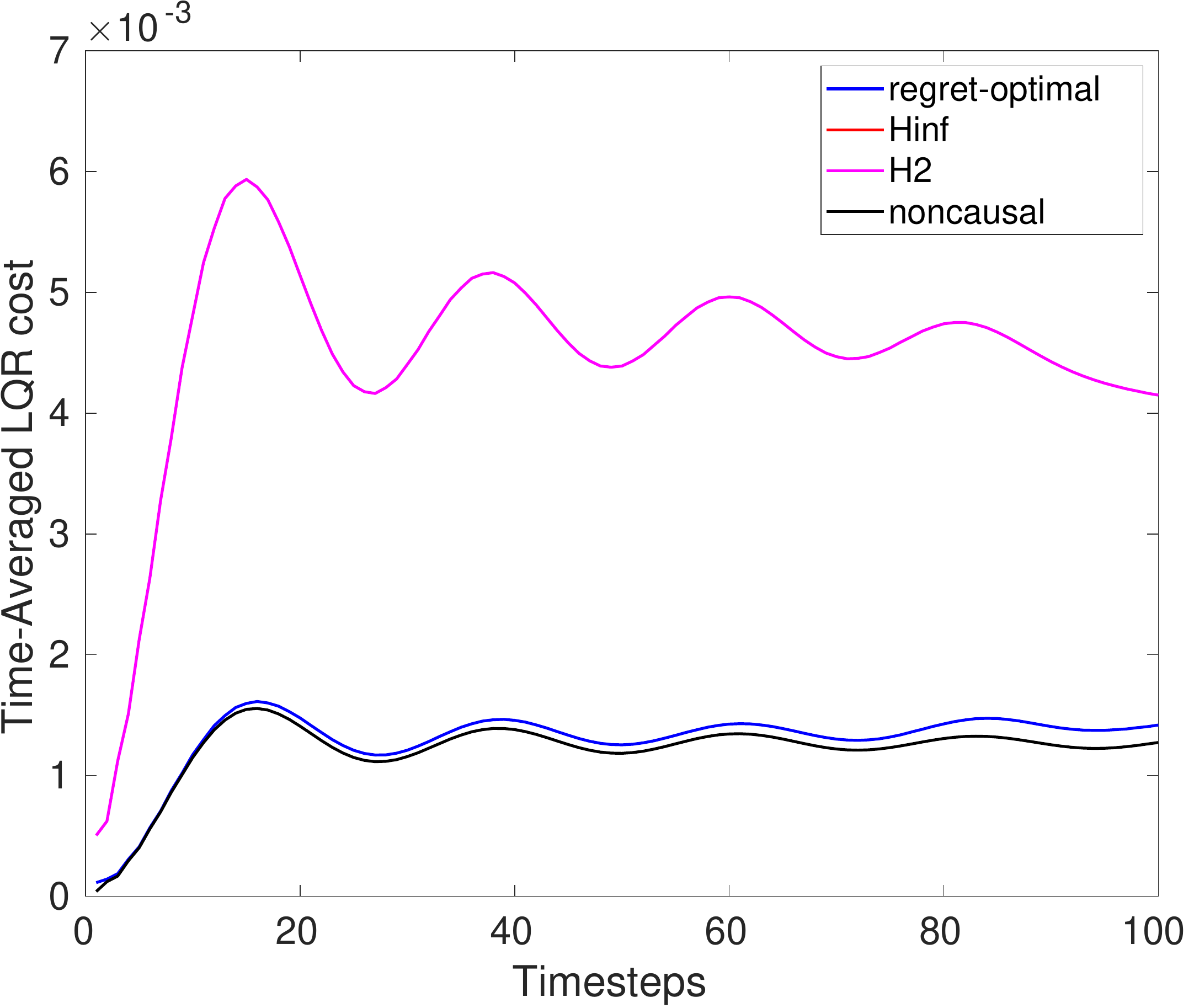}
\caption{Relative performance of LQR controllers with sinusoidal noise $w(k\Delta) = \sin{(30k\Delta)}$.}
\label{sinusoidal-fig}
\end{figure}

We benchmark our regret-optimal controller in the classic inverted pendulum model. This system has two scalar states, $\theta$ and $\dot{\theta}$, representing angular position and angular velocity, respectively, and a single scalar control input $u$. The states evolve according to the nonlinear evolution equation 

\begin{equation*}
    \frac{d}{dt} \begin{bmatrix} \theta \\ \dot{\theta} \end{bmatrix} = \begin{bmatrix} \dot{\theta} \\ \frac{mg\ell}{J}\sin{\theta} + \frac{\ell}{J}u\cos{\theta} + \frac{\ell}{J}w\cos{\theta} \end{bmatrix},
\end{equation*}
where $w$ is an external disturbance, and $m, g, \ell,$ and $J$ are physical parameters. Although these dynamics are nonlinear, we can benchmark the regret-optimal controller against the $H_2$-optimal, $H_{\infty}$-optimal, and clairvoyant noncausal controllers using Model Predictive Control (MPC). In the MPC framework, we iteratively linearize the model dynamics around the current state, compute the optimal control signal in the linearized system, and then update the state in the original nonlinear system using this control signal. In our experiments we take $Q, R = I$ and assume that units are scaled so that all physical parameters are 1. We set the discretization parameter $\Delta = 0.1$ and sample the costs at times $t = k\Delta$ as $k$ ranges from 1 to 100. We initialize both the angular position and the angular velocity to zero. In our first experiment, the disturbance $w$ is drawn from a standard Gaussian distribution (Figure \ref{gaussian-fig}). The $H_2$-optimal controller incurs the lowest cost; this is unsurprising, since the $H_2$ controller is designed to minimize the expected cost when the disturbances are stochastic. We note that the regret-optimal controller closely tracks the performance of the $H_2$ controller, and significantly outperforms the $H_{\infty}$ controller. In our second experiment, $w$ is a sawtooth signal (Figure \ref{sawtooth-fig}); the regret-optimal controller achieves an order of magnitude less cost than the $H_{\infty}$-optimal controller and also outperforms the $H_2$ controller. In our third experiment, $w$ is a sinusoidal signal. We select $w(k\Delta) = \sin{(10k\Delta)}$ (Figure \ref{sinusoidal-fig-2}) and $w(k\Delta) = \sin{(30k\Delta)}$ (Figure \ref{sinusoidal-fig}). In both settings, the regret-optimal controller closely tracks the performance of the clairvoyant noncausal controller and achieves two orders of magnitude better performance than the $H_{\infty}$ controller. 

%experiment 2: ws = sawtooth(2*(1:100));
%experiment 3: ws = sin(3*(1:100));

\begin{figure}[h]
\centering
\includegraphics[width=0.9\columnwidth]{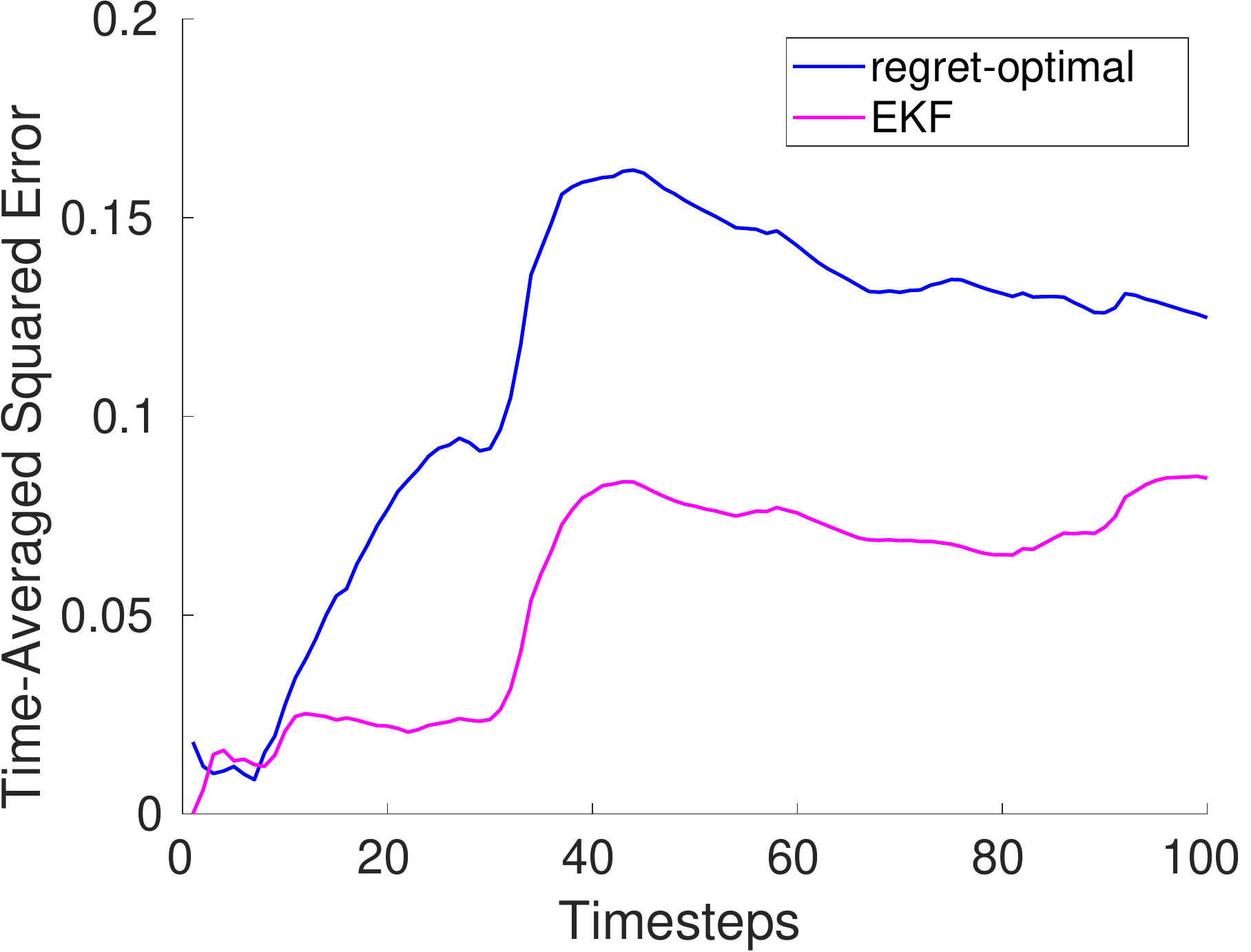}
\caption{Relative performance of regret-optimal filter and the Extended Kalman Filter (EKF) with Gaussian noise.}
\label{gaussian-estimation-fig}
\end{figure}

\begin{figure}[h]
\centering
\includegraphics[width=0.9\columnwidth]{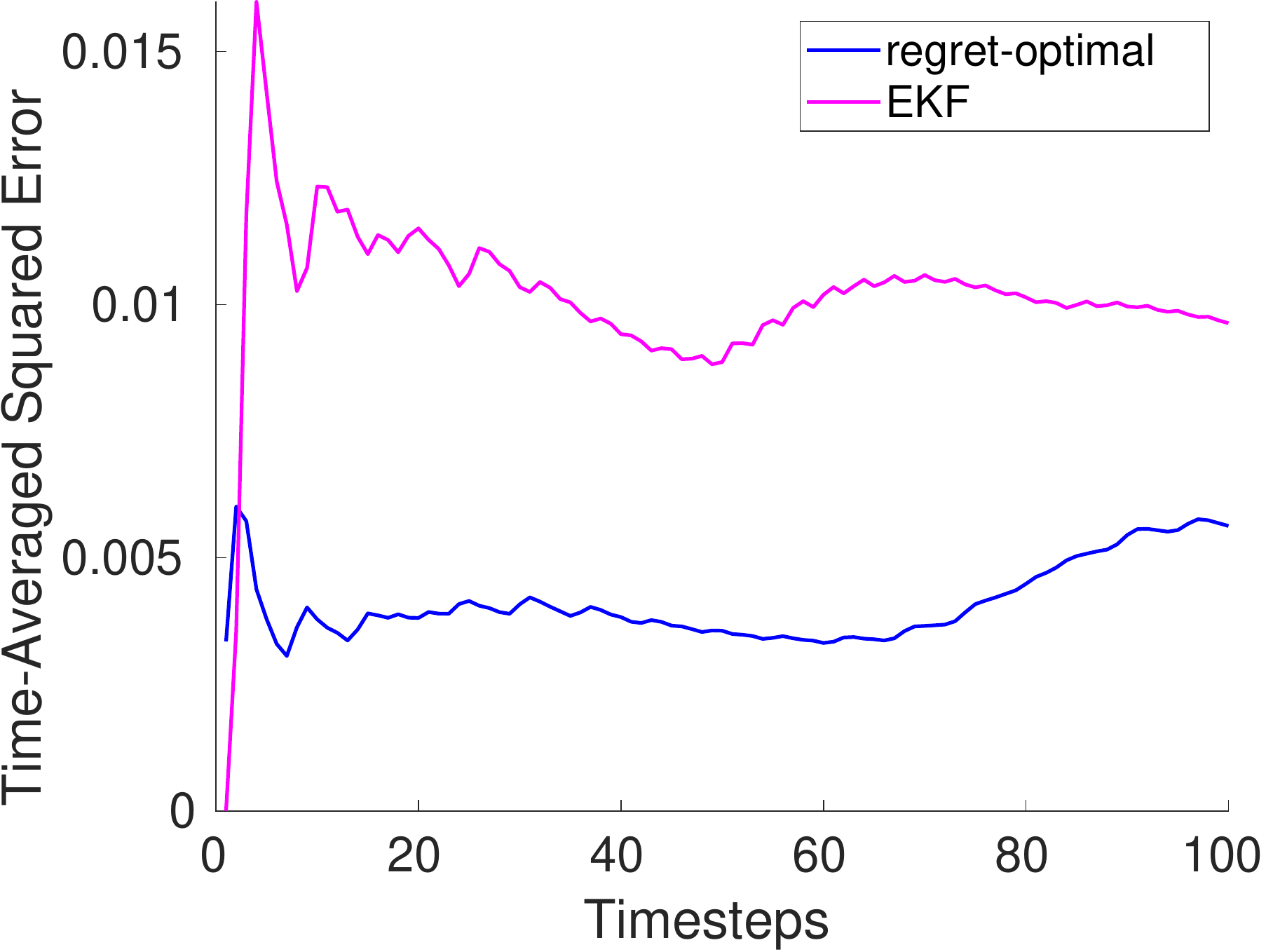}
\caption{Relative performance of regret-optimal filter and the Extended Kalman Filter (EKF) with sinusoidal noise $u(k\Delta) = \sin{(10k\Delta)}, v(k\Delta) = \cos{(10k\Delta)}$.}
\label{sinusoidal-estimation-fig}
\end{figure}

\begin{figure}[h]
\centering
\includegraphics[width=0.9\columnwidth]{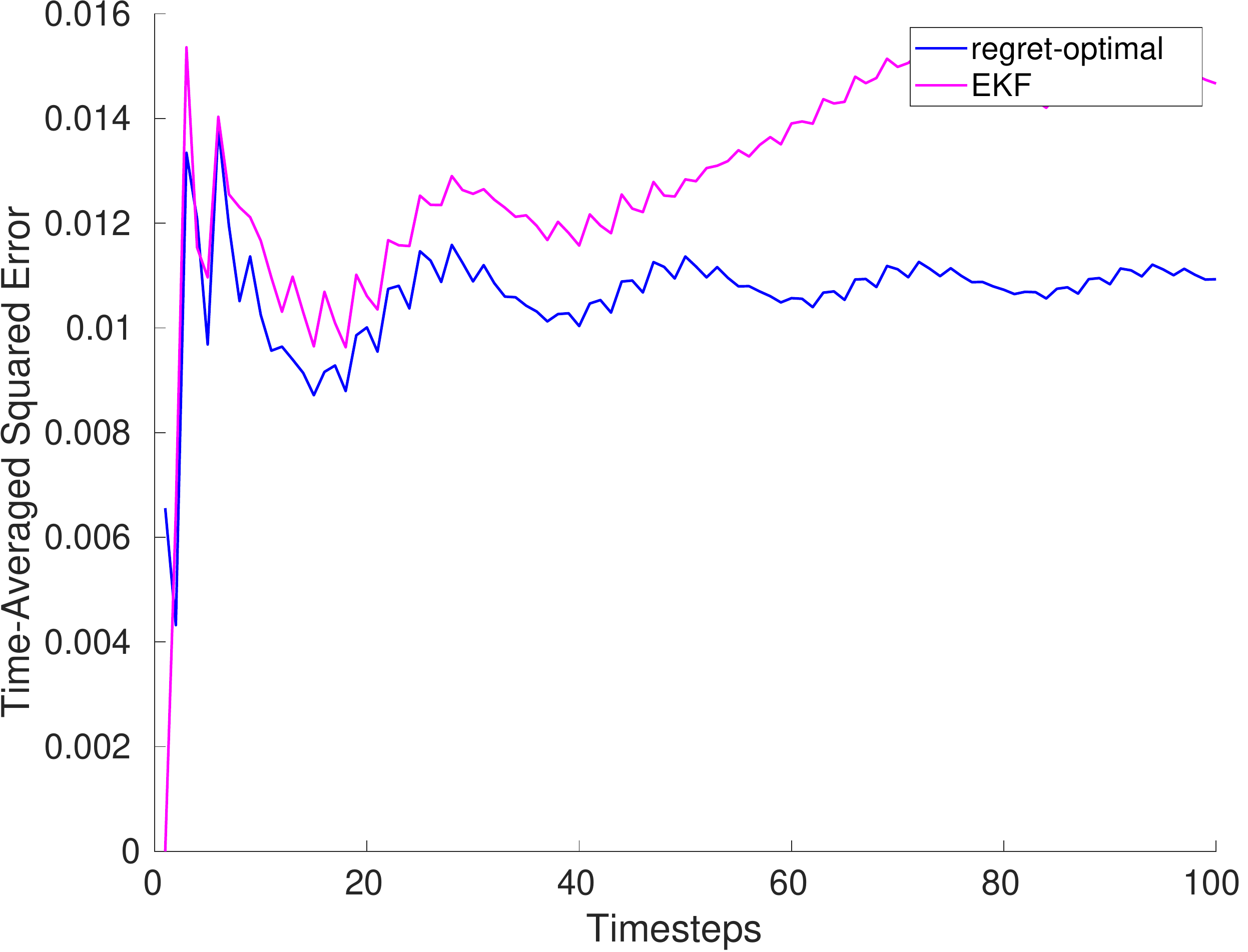}
\caption{Relative performance of regret-optimal filter and the Extended Kalman Filter (EKF) with sinusoidal noise $u(k\Delta) = \sin{(10k\Delta)} + 2\cos{(30k\Delta)}, v(k\Delta) = \cos{(10k\Delta)} + 0.5\sin{(10k\Delta)}$.}
\label{sinusoidal-estimation-fig-2}
\end{figure}

We next benchmark the regret-optimal filter. We consider frequency modulation, a classic setting in communications theory where a message is passed through an integrator to phase modulate a carrier signal. In this problem, there are two states, $\lambda(t)$ and $\theta(t)$. The dynamics are linear and time-invariant, and are given by $$\frac{d}{dt} \begin{bmatrix} \lambda(t) \\ \theta(t) \end{bmatrix} = \begin{bmatrix} -1/\beta & 0 \\ 1 & 0 \end{bmatrix}\begin{bmatrix} \lambda(t) \\ \theta(t) \end{bmatrix} + \begin{bmatrix} 1 \\ 0 \end{bmatrix}u(t).$$ The observations are nonlinear in the state: $$ y(t) = \sqrt{2}\sin{[\omega_ct + \theta(t)]} + v(t).$$ While the observation model is nonlinear, we can apply the regret-optimal filter by iteratively linearizing the dynamics around the current estimate, computing the regret-optimal filter in the linearized system, and forming a new estimate using this filter; this is the same approach used in the classic Extended Kalman filter (EKF) algorithm (see \cite{kailath2000linear} for background on the EKF). We assume that both $\lambda$ and $\theta$ are initialized to zero and set $\beta, \omega_c = 1$. We take the disturbance covariances $Q, R = I$ in our EKF implementation. As in our control experiments, we set the discretization parameter $\Delta = 0.1$ and sample the costs at times $t = k\Delta$ as $k$ ranges from 1 to 100. We benchmark the regret-optimal filter against the EKF across a variety of input disturbances. In our first experiment, the disturbances $u, v$ are both selected i.i.d from a standard Gaussian distribution (Figure \ref{gaussian-estimation-fig}). We see that the EKF outperforms the regret-optimal filter; this is unsurprising, since the EKF is tuned for stochastic noise. In our second set of experiments, we generate sinusoidal disturbances.  First, we take $u(k\Delta) = \sin{(10k\Delta)}, v(k\Delta) = \cos{(10k\Delta)}$; we see that the regret-optimal filter incurs roughly half the squared estimation error of the EKF (Figure \ref{sinusoidal-estimation-fig}). We next consider a more complex set of input disturbances, namely $u(k\Delta) = \sin{(10k\Delta)} + 2\cos{(30k\Delta)}, v(k\Delta) = \cos{(10k\Delta)} + 0.5\sin{(10k\Delta)}$, and obtain similar results (Figure \ref{sinusoidal-estimation-fig-2}).

Together, our experiments show that by minimizing regret against the noncausal estimators and controllers, our regret-optimal algorithms are able to adapt to many different kinds of input disturbances.

\section{Conclusion}
We propose regret against a clairvoyant noncausal policy as a criterion for estimator and controller design, and show that regret-optimal estimators and controllers can be found by extending $H_{\infty}$ estimation and control to minimize regret instead of just cost. We give a complete characterization of regret-optimal estimators and controllers in state-space form, allowing efficient implementations whose computational cost scales linearly in the time horizon. We also give tight bounds on the regret incurred by our algorithms in terms of the energy of the disturbances. Numerical benchmarks in nonlinear systems show that our regret-optimal algorithms are able to adapt to many different kinds of disturbances and can often outperform standard $H_2$-optimal and $H_{\infty}$-optimal algorithms.

We identify several promising directions for future research. First, it is natural to consider minimizing regret against a controller which in each timestep has access to $k$ predictions of future disturbances, instead of the full sequence of disturbances; such a controller is more easily implemented in real-world systems, and the resulting regret-minimizing controller may demonstrate better performance. Second, it would be interesting to use our regret-minimization techniques to design decentralized controllers with access to local information which compete against a centralized controller with global information; such a distributed regret-optimal controller may have applications in network optimization and control. 

\newpage

\section{Appendix}

\subsection{Proof of Theorem \ref{regret-optimal-estimation-thm}} \label{regret-optimal-estimation-sec}

A state-space model for $\mathcal{H}$ is given by $$x_{t+1} = A_tx_t + B_t u_t, \hspace{5mm} y_t = C_tx_t,$$
so a state-space model for $\mathcal{H}^*$ is $$x_{t-1} = A_t^{\top}x_t + C_t^{\top} y_t, \hspace{5mm} u_t = B^{\top}_tx_t.$$
Let $w$ be such that $\expect[w_tw_t^{\top}] = I$ and $\expect[w_t^{\top}y_t] = 0$, and define $u = \mathcal{H}^*y + w$. Notice that $\expect[u u^*] = I + \mathcal{H}^*\mathcal{H}$. Suppose we can find an causal operator $\Delta_1$ such that $u = \Delta_1^* e$ where $e$ is a random variable with mean zero such that $\expect[ee^*] = I$. Then necessarily $\Delta_1^* \Delta_1 = I + \mathcal{H}^*\mathcal{H}$. Using the backwards-time Kalman filter, we obtain a state-space model for $\Delta_1^*$: $$x_{t-1} = A_t^{\top}x_t + K_t\Sigma_t^{1/2} e_t, \hspace{5mm} u_t = B_t^{\top}x_t + \Sigma_t^{1/2}e_t,$$ where $K_t = A_t^{\top}P_{t}B_t \Sigma_t^{-1}$, $\Sigma_t = I + B_t^{\top}P_{t}B_t,$ and $P_{t}$ is the solution of the recurrence $$P_{t-1} = A_t^{\top}P_{t}A_t + C_t^{\top}C_t - K_t \Sigma_t K_t^{\top}$$ where we initialize $P_{T-1} = 0$.
It follows that a state-space model for $\Delta_1$ is $$x_{t+1} = A_tx_t + B_t u_t, \hspace{5mm} e_t = \Sigma_t^{1/2}K_t^{\top}x_t + \Sigma_t^{1/2}u_t.$$ Exchanging inputs and outputs, we see that a state-space model for $\Delta_1^{-1}$ is $$x_{t+1} = (A_t - B_tK_t^{\top})x_t + B_t \Sigma_t^{-1/2}e_t,$$ $$ u_t = -K_t^{\top}x_t +  \Sigma_t^{-1/2}e_t.$$ A state-space model for $\mathcal{L}$ is given by $$x_{t+1} = A_tx_t + B_t u_t, \hspace{5mm} s_t = L_t x_t.$$ After some simplification, we see that a state-space model for $\mathcal{L}\Delta_1^{-1}$ is $$y_{t+1} = (A_t - B_tK_t^{\top})y_t + B_t \Sigma_t^{-1/2}e_t, \hspace{5mm} s_t = L_t y_t.$$ 

\noindent We now turn to the problem of factoring $I + \gamma^{-2}\mathcal{L}(I + \mathcal{H}^*\mathcal{H})^{-1}\mathcal{L}^*$ as $\mathcal{S}^*\mathcal{S}$, where $\mathcal{S}$ is causal. We have $$I + \gamma^{-2}\mathcal{L}(I + \mathcal{H}^*\mathcal{H})^{-1}\mathcal{L}^* = I + \gamma^{-2}(\mathcal{L}\Delta_1^{-1})(\mathcal{L}\Delta_1^{-1})^*.$$ Notice that $(\mathcal{L}\Delta_1^{-1})(\mathcal{L}\Delta_1^{-1})^*$ is the product of a causal operator and an anticausal operator, whereas in the factorization $\mathcal{S}^*\mathcal{S}$ the order of the causal and anticausal factors is reversed, so we cannot directly apply the Kalman filter technique we used to factor $I + \mathcal{H}^*\mathcal{H}$ as $\Delta_1^* \Delta_1 $. Consider the model $$y_{t+1} = (A_t - B_tK_t^{\top})y_t + B_t \Sigma_t^{-1/2}e_t, \hspace{5mm} s_t = \gamma^{-1}L_t y_t + f_t,$$ where $f_t$ is a zero-mean r.v. such that $\expect[f_tf_t^{\top}] = I, \expect[f_te_t^{\top}] = 0$. Notice that $$\expect[ss^*] = I + \gamma^{-2}\mathcal{L}(I + \mathcal{H}^*\mathcal{H})^{-1}\mathcal{L}^*. $$ 
Applying Theorem 9.8.1 in \cite{kailath2000linear}, we see that a state-space model for $\mathcal{S}^*$ is given by $$y^b_{t-1} = A^b_{t}y^b_t + K^b_t(\Sigma_t^b)^{1/2} f_t^b, \hspace{5mm} s_t = \gamma^{-1}L_ty_t^b + (\Sigma_t^b)^{1/2} f_t^b, $$ where $A_t^b$ is any solution to the equation $$A_t^b \Pi_t = \Pi_{t-1}(A_{t-1} - B_{t-1}K_{1, t-1}^{\top})^{\top}$$ and we define $$K_t^b = \gamma^{-1} A_t^bP_t^bL_t^{\top}(\Sigma^b_t)^{-1},$$ $$\Sigma_t^b = I + \gamma^{-2}L_t P_t^b L_t^{\top},$$ $$Q_t^b = \Pi_{t-1} - A_t^b \Pi_t (A_t^b)^{\top}.$$ We define the state-covariance matrix $\Pi_t$ to be the solution of the recurrence $$\Pi_{t+1} =  (A_t - B_tK_t^{\top})\Pi_t(A_t - B_tK_t^{\top})^{\top} + B_t \Sigma_t^{-1}B_t^{\top},$$ and define $P_t^b$ to be the solution to the Riccati recursion $$P_{t-1}^b = A_t^bP_t(A_t^b)^{\top} + Q_t^b - K^b_t \Sigma_t^b (K_t^b)^{\top}, $$ where we initialize $P_{T-1} = \Pi_{T-1}$. It follows that a state-space model for $\mathcal{S}$ is $$z_{t+1} = (A^b_{t})^{\top}z_t + \gamma^{-1}L_t^{\top}s_t, \hspace{3mm} f_t = (\Sigma_t^b)^{1/2}(K^b_t)^{\top}z_t + (\Sigma_t^b)^{1/2}s_t,$$ and a state-space model for $\mathcal{S}^{-1}$ is $$z_{t+1} = \left((A^b_{t})^{\top} - \gamma^{-1}L_t^{\top} (K^b_t)^{\top} \right)z_t + \gamma^{-1}L_t^{\top}(\Sigma_t^b)^{-1/2}f_t,$$ $$ s_t = -(K^b_t)^{\top}z_t +(\Sigma_t^b)^{-1/2}f_t.$$

Recall that $\mathcal{T}$ is the unique causal operator such that 
\begin{eqnarray*}
\mathcal{T}\mathcal{T}^* &=& \gamma^2 (\mathcal{S}^*\mathcal{S})^{-1} + \mathcal{L}(I + \mathcal{H}^*\mathcal{H})^{-1}\mathcal{L}^* \\
&=& (\gamma\mathcal{S}^{-1})(\gamma\mathcal{S}^{-1})^* + (\mathcal{L}\Delta_1^{-1})(\mathcal{L}\Delta_1^{-1})^*.
\end{eqnarray*}
Putting the pieces together, we see that a state-space model for $\mathcal{T}$ is given by 
$$y_{t+1} = (A_t - B_tK_t^{\top})y_t + B_t \Sigma_t^{-1/2}e_t, $$ $$ z_{t+1} = \left((A^b_{t})^{\top} - \gamma^{-1}L_t^{\top} (K^b_t)^{\top} \right)z_t + \gamma^{-1}L_t^{\top}(\Sigma_t^b)^{-1/2}f_t, $$ 
$$ s_t = L_ty_t - \gamma(K^b_t)^{\top}z_t + \gamma(\Sigma_t^b)^{-1/2}f_t .$$
Define 
\begin{align}
\eta_t &= \begin{bmatrix} y_t \\z_t \end{bmatrix} \nonumber \\
\tilde{A}_t &= \begin{bmatrix} A_t - B_tK_t^{\top} & 0 \\ 0 & (A^b_{t})^{\top} - \gamma^{-1}L_t^{\top} (K^b_t)^{\top} \end{bmatrix}, \nonumber \\
\tilde{B}_t &=  \begin{bmatrix} B_t \Sigma_t^{-1/2} & 0 \\ 0 & \gamma^{-1}L_t^{\top}(\Sigma_t^b)^{-1/2} \end{bmatrix} \nonumber  \\
\tilde{C}_t &= \begin{bmatrix} L_t & - \gamma(K^b_t)^{\top} \end{bmatrix}. \label{tilde-matrices-estimation}
\end{align}
With this notation, our state-space model for $\mathcal{T}$ takes a compact form: 
$$\eta_{t+1} = \tilde{A}_t \eta_t + \tilde{B} \begin{bmatrix} e_t \\ f_t \end{bmatrix}, \hspace{3mm} s_t = \tilde{C}_t \eta_t +  \begin{bmatrix} 0 & \gamma(\Sigma_t^b)^{-1/2} \end{bmatrix} \begin{bmatrix} e_t \\ f_t \end{bmatrix}.$$

Applying the Kalman filter once again, we obtain a whitened model for $\mathcal{T}$: $$\eta_{t+1} = \tilde{A}_t \eta_t +  \tilde{K}_t \tilde{\Sigma}^{1/2}_t g_t, \hspace{5mm} s_t = \tilde{C}_t \eta_t + \tilde{\Sigma}^{1/2}_t  g_t,$$ where 
\begin{align}
\tilde{K}_t &= \left(\tilde{A}_t \tilde{P}_t \tilde{C}_t^{\top} + \tilde{B}_t \begin{bmatrix} 0 \\ \gamma(\Sigma_t^b)^{-1/2} \end{bmatrix} \right)\tilde{\Sigma}_t^{-1} \nonumber \\ \tilde{\Sigma}_t &= \gamma^2(\Sigma_t^b)^{-1} + \tilde{C}_t\tilde{P}_t\tilde{C}_t^{\top}, \label{tilde-matrices-estimation2}
\end{align}
and we define $\tilde{P}_t$ to be the solution of the recurrence $$\tilde{P}_{t+1} = \tilde{A}_t\tilde{P}_t\tilde{A}_t^{\top} + \tilde{B}_t\tilde{B}_t^{\top} - \tilde{K}_t \tilde{\Sigma}_t \tilde{K}_t^{\top}$$ and initialize $\tilde{P}_0 = 0$.  Exchanging inputs and outputs once again, we see that a model for $\mathcal{T}^{-1}$ is $$\eta_t = (\tilde{A}_t - \tilde{K}_t \tilde{C}_t)\eta_t +  \tilde{K}_t s_t, \hspace{5mm} g_t = -\tilde{\Sigma}_t^{-1/2} \tilde{C}_t \eta_t + \tilde{\Sigma}^{-1/2}_t  s_t.$$

Recall that the regret-optimal controller is $\mathcal{K} = \mathcal{T}\mathcal{K}_{\infty}$, where $\mathcal{K}_{\infty}$ is the $H_{\infty}$-estimator at level $\gamma = 1$ of the variable $g = \mathcal{T}^{-1}\mathcal{L}u$ given observations $y = \mathcal{H}u + v$. A state-space model for $\mathcal{T}^{-1}\mathcal{L}$ is $$\begin{bmatrix} x_{t+1} \\ \eta_{t+1} \end{bmatrix} = \hat{A}_t \begin{bmatrix} x_t \\ \eta_t \end{bmatrix} + \hat{B}_tu_t, \hspace{5mm} g_t =   \hat{L}_t \begin{bmatrix} x_t \\ \eta_t \end{bmatrix},$$ where we define $$\hat{A}_t = \begin{bmatrix} A_t & 0  \\ \tilde{K}_t L_t & \tilde{A}_t - \tilde{K}_t \tilde{C}_t \end{bmatrix},$$ $$ \hat{B}_t = \begin{bmatrix}  B_t \\ 0 \end{bmatrix},  \hspace{5mm} \hat{L}_t = \begin{bmatrix} \tilde{\Sigma}_t^{-1/2}L_t & -\tilde{\Sigma}_t^{-1/2}\tilde{C}_t \end{bmatrix}. $$
Recall that we receive the observations $$y_t = C_t x_t + v_t = \hat{C}_t \begin{bmatrix} x_t \\ \eta_t \end{bmatrix} + v_t,$$ where we define $$\hat{C}_t = \begin{bmatrix} C_t & 0 \end{bmatrix}.$$

Feeding this into the state-space description for a causal $H_{\infty}$ estimator at level $\gamma = 1$ given in Theorem \ref{hinf-suboptimal-estimation-thm}, we see that a model for $\mathcal{K}_{\infty}$ is given by 
$$\hat{g}_t = \hat{L}_t \begin{bmatrix} \hat{x}_{t \mid t} \\ \hat{\eta}_{t \mid t} \end{bmatrix},$$ where
$$\begin{bmatrix} \hat{x}_{t+1 \mid t+1} \\ \hat{\eta}_{t + 1\mid t + 1} \end{bmatrix} = \hat{A}_t \begin{bmatrix} \hat{x}_{t\mid t} \\ \hat{\eta}_{t \mid t} \end{bmatrix} + \hat{K}_{t+1} (y_{t+1} - C_{t+1}A_t \hat{x}_{t\mid t} ), $$ where we initialize $\hat{x}_{0 \mid 0} = 0, \hat{\eta}_{0 \mid 0} = 0$ and we define $$\hat{K}_t = \hat{P}_t\hat{C}_t^{\top}(I + \hat{C}_t\hat{P}_t\hat{C}_t^{\top})^{-1},$$ $$\hat{\Sigma}_t = \begin{bmatrix} I & 0 \\ 0 & -I \end{bmatrix} + \begin{bmatrix} \hat{C}_t \\ \hat{L}_t \end{bmatrix}  \hat{P}_t \begin{bmatrix} \hat{C}_t^{\top} & \hat{L}_t^{\top} \end{bmatrix},$$
and $\hat{P}_t$ is the solution of the Riccati recursion $$\hat{P}_{t+1} = \hat{A}_t \hat{P}_t \hat{A}_t^{\top} + \hat{B}_t\hat{B}_t^{\top} -  \hat{A}_t \hat{P}_t \begin{bmatrix} \hat{C}_t^{\top} & \hat{L}_t^{\top} \end{bmatrix}\hat{\Sigma}^{-1}_t \begin{bmatrix} \hat{C}_t \\ \hat{L}_t \end{bmatrix} \hat{P}_t  \hat{A}_t^{\top},$$
where we initialize $\hat{P}_0 = 0$.
Recall that $\mathcal{K} = \mathcal{T}\mathcal{K}_{\infty}$. Plugging the output of our state-space model for $\mathcal{T}$ into the model for $\mathcal{K}_{\infty}$ and simplifying, we see that a model for $\mathcal{K}$ is
  $$ \hat{s}_t = L_t \hat{x}_{t \mid t},$$
where $\hat{x}_{t \mid t}$ is defined in the model for $\mathcal{K}_{\infty}$.
It is easy to extend this result to the strictly causal setting, where in each timestep the estimator outputs an estimate $\hat{s}_t$ using only the observations $y_0, \ldots y_{t-1}$; we simply use a strictly causal model for $\mathcal{K}_{\infty}$ in place of the causal model (see Theorem 4.2.2 in \cite{hassibi1999indefinite} for such a model). 
The regret bounds appearing in Theorem \ref{regret-optimal-estimation-thm} are immediate; by definition, the regret-optimal estimator at level $\gamma$ has regret at most $\gamma^2 (\|u\|_2^2 + \|w\|_2^2)$. To find the optimal value of $\gamma$, we minimize $\gamma$ subject to the constraints described in Theorem \ref{hinf-suboptimal-estimation-thm}.
$\blacksquare$

\subsection{Proof of Theorem \ref{regret-optimal-controller-thm}}
A state-space model for $\mathcal{F}$ is given by $$\xi_{t+1} = A_t \xi_t + B_{u, t} u_t, \hspace{5mm} s_t = Q_t^{1/2} \xi_t.$$ Let $v_t$ be a zero-mean r.v. such that $\expect[v_t v_t^{\top}] = I$ and $\expect[u_t v_t^{\top}] = 0$. Define $y = \mathcal{F} u + v$; notice that $\expect[y y^{\top}] = I + \mathcal{F}\mathcal{F}^*$. Suppose we can find a causal matrix $\Delta_1$ such that $y = \Delta_1 e$ where $e$ is a zero-mean random variable such that $\expect[e e^{\top}] = I$. Then $\expect[y y^{\top}] = \Delta_1 \Delta_1^*$, so $I + \mathcal{F}\mathcal{F}^* = \Delta_1 \Delta_1^*$. 

Using the Kalman filter (as described in Theorem 9.2.1 in \cite{kailath2000linear}), we obtain a state-space model for $\Delta_1$:
\begin{equation} \label{first-factorization}
\hat{\xi}_{t+1} = A_t \hat{\xi}_t + K_{t} \Sigma^{1/2}_{t} e_t, \hspace{5mm} y_t = Q_t^{1/2} \hat{\xi}_t +  \Sigma_t^{1/2}e_t,
\end{equation}
where we define $K_{t} = A_t P_t Q_t^{1/2} \Sigma^{-1}$ and $\Sigma_t = I + Q_t^{1/2}P_t Q_t^{1/2}$ and $P_t$ is defined recursively as $$P_{t+1} = A_tP_tA_t^{\top} + B_{u, t}B_{u, t}^{\top} - K_t \Sigma_tK_t^{\top}$$ and $P_0 = 0$. Exchanging inputs and outputs, we see that a state-space model for $\Delta_1^{-1}$ is $$\hat{\xi}_{t+1} = A_t\hat{\xi}_t + K_t(y_t - Q_t^{1/2}\hat{\xi}_t), \hspace{5mm} e_t = \Sigma_t^{-1/2}(y_t - Q_t^{1/2} \hat{\xi}_t).$$ We have factored $I + \mathcal{F}\mathcal{F}^*$ as $\Delta_1 \Delta_1^*$, so $\gamma^2 I +  \mathcal{G}^*(I + \mathcal{F}\mathcal{F}^*)^{-1}\mathcal{G} = \gamma^2 I + (\Delta_1^{-1} \mathcal{G})^{\top} (\Delta_1^{-1} \mathcal{G})$. Notice that $\Delta_1^{-1}\mathcal{G}$ is strictly causal, since $\Delta_1^{-1}$ is causal and $\mathcal{G}$ is strictly causal. A state-space model for $\mathcal{G}$ is $$\eta_{t+1} = A_t \eta_t + B_{w, t} w_t, \hspace{5mm} s_t = Q_t^{1/2} \eta_t.$$ 

\noindent Equating $s$ and $y$, we see that a state-space model for $\Delta_1^{-1}\mathcal{G}$ is  $$\begin{bmatrix} \hat{\xi}_{t+1} \\ \eta_{t+1} \end{bmatrix} = \begin{bmatrix} \tilde{A}_t & K_tQ_t^{1/2}  \\ 0 & A_t  \end{bmatrix} \begin{bmatrix} \hat{\xi}_{t} \\ \eta_{t} \end{bmatrix} + \begin{bmatrix} 0 \\ B_{w, t}  \end{bmatrix} w_t, $$ $$e_t =  \Sigma_t^{-1/2}Q_t^{1/2}( \eta_t - \hat{\xi}_t), $$
where we defined $\tilde{A}_t = A_t - K_t Q_t^{1/2}$. Setting $\nu_t = \eta_t - \hat{\xi}_t$ and simplifying, we see that a minimal representation for a state-space model for $\Delta_1^{-1}\mathcal{G}$ is  

\begin{equation} \label{intermediate-ss}
\nu_{t+1} = \tilde{A}_t\nu_t + B_{w, t}w_t, \hspace{5mm} e_t = \Sigma_t^{-1/2} Q_t^{1/2} \nu_t,
\end{equation}
It follows that a state-space model for $(\Delta_1^{-1}\mathcal{G})^{\top}$ is $$\nu_{t-1} = \tilde{A}_t^{\top}\nu_t + Q_t^{1/2} \Sigma_t^{-1/2}  w_t, \hspace{5mm} e_t =  B_{w, t}^{\top} \nu_t.$$ 

Recall that our original goal was to obtain a factorization  $\gamma^2 I +  \mathcal{G}^*(I + \mathcal{F}\mathcal{F}^*)^{-1}\mathcal{G} = \Delta_2^{\top}\Delta_2$, where $\Delta$ is causal. Let $z = (\Delta_1^{-1}\mathcal{G})^{\top} a + b$, where $a$ and $b$ are zero-mean random variables such that $\expect[a a^{\top}] = I, \expect[a b^{\top}] = 0, $ and $\expect[b b^{\top}] = \gamma^2 I$. Suppose that we can find an causal matrix $\Delta_2$ such that $z = \Delta_2^* f$, where $f$ is a zero-mean random variable such that $\expect[f f^{\top}] = I$. Notice that $\expect[z z^{\top}] = \gamma^2 I + (\Delta_1^{-1} \mathcal{G})^{\top} (\Delta_1^{-1} \mathcal{G}) =  \gamma^2 I +  \mathcal{G}^*(I + \mathcal{F}\mathcal{F}^*)^{-1}\mathcal{G}$; on the other hand $\expect[zz^{\top}] = \Delta_2^*\Delta_2$ as desired.

A backwards-time state-space model for $z$ is given by $$ \nu_{t-1} = \tilde{A}_t^{\top}\nu_t - Q_t^{1/2} \Sigma_t^{-1/2} a_t, $$ $$ z_{t} = B_{w, t}^{\top} \nu_t + b_t.$$ Using the (backwards time) Kalman filter, we see that a state-space model for $\Delta_2^*$ is given by $$\nu_{t-1} =  \tilde{A}_t^{\top}\nu_t +K_t^b (\Sigma_t^b)^{1/2}f_t, $$ $$ z_t = B_{w, t}^{\top}\nu_t  + (\Sigma_t^b)^{1/2}f_t,$$ where we define $K_t^b = \tilde{A}_t^{\top}P_t^b B_{w, t} (\Sigma_t^b)^{-1} $  and $\Sigma_t^b = \gamma^2 I + B^{\top}_{w, t} P_t^b B_{w, t}$, and $P_t^b$ is the solution to the backwards Riccati recursion 
$$P_{t-1}^b =  \tilde{A}_t^{\top}P_{t}^b \tilde{A}_t + Q_t^{1/2}\Sigma_t^{-1}Q_t^{1/2} - K_t^b \Sigma_t^b (K_t^b)^{\top}, $$ and $P_T^b = 0$.  It follows that a state-space model for $\Delta_2$ is
\begin{align} \label{model-for-delta2}
\nu_{t+1} &=  \tilde{A}_t\nu_t + B_{w, t} f_t, \\
z_t &= (\Sigma_t^b)^{1/2} (K_t^b)^{\top}\nu_t  +   (\Sigma_t^b)^{1/2}f_t. \nonumber
\end{align}
Therefore a state-space model for $\Delta_2^{-1}$ is $$ \nu_{t+1} =  (\tilde{A}_t - B_{w, t}(K_t^b)^{\top})\nu_t + B_{w, t}(\Sigma_t^b)^{-1/2}z_t,$$ $$ f_t = -(K_t^b)^{\top}\nu_t + (\Sigma_t^b)^{-1/2}z_t .$$ 
 
 \noindent Recall that a state-space model for $\mathcal{G}$ is  $$\eta_{t+1} = A_t \eta_t + B_{w, t} w_t, \hspace{5mm} s_t = Q_t^{1/2} \eta_t.$$ Equating $f_t$ and $w_t$, we see that a state-space model for $\mathcal{G}\Delta_2^{-1}$ is $$\nu_{t+1} =  (\tilde{A}_t - B_{w, t}(K_t^b)^{\top})\nu_t + B_{w, t}(\Sigma_t^b)^{-1/2}z_t,$$ $$ \eta_{t+1} = A_t \eta_t - B_{w, t}(K_t^b)^{\top}\nu_t + B_{w, t} (\Sigma_t^b)^{-1/2}z_t, $$ $$ s_t = Q_t^{1/2} \eta_t.$$
Recall that a state-space model for $\mathcal{F}$ is  $$\xi_{t+1} = A_t\psi_t + B_{u, t}u_t, \hspace{5mm} s_t = Q_t^{1/2}\xi_t.$$ Letting $\zeta_t = \eta_t + \xi_t$, we see that a state-space model for the overall system is
\begin{align} \label{regret-final-dynamics}
\begin{bmatrix} \zeta_{t+1} \\ \nu_{t+1} \end{bmatrix} &= \begin{bmatrix} A_t & - B_{w, t}(K_t^b)^{\top} \\ 0 &  \tilde{A}_t - B_{w, t}(K_t^b)^{\top} \end{bmatrix} \begin{bmatrix} \zeta_t \\ \nu_t \end{bmatrix} \\ \nonumber
 &+ \begin{bmatrix} B_{u, t} \\ 0 \end{bmatrix} u_t +  \begin{bmatrix} B_{w, t} (\Sigma_t^b)^{-1/2} \\ B_{w, t} (\Sigma_t^b)^{-1/2} \end{bmatrix} z_t, \\
 s_t &=  \begin{bmatrix} Q_t^{1/2} & 0 \end{bmatrix} \begin{bmatrix} \zeta_t \\ \nu_t \end{bmatrix}. \nonumber
\end{align}
To derive the regret-suboptimal controller, we can plug this state-space model into the formula for the  $H_{\infty}$ controller at level $\gamma = 1$ given in Theorem \ref{hinf-suboptimal-controller-thm}. We see that the regret-optimal controller is given by 
\begin{equation} \label{regret-optimal-controller-description-proof}
 u_t = -\hat{H}_t^{-1} \hat{B}_{u, t}^{\top} \hat{P}_{t+1} \left(\hat{A}_t \begin{bmatrix} \zeta_t \\ \nu_t \end{bmatrix} + \hat{B}_{w, t} z_t \right), 
\end{equation}
 where we define 
$$\hat{A}_t = \begin{bmatrix} A_t & - B_{w, t}(K_t^b)^{\top} \\ 0 &  \tilde{A}_t - B_{w, t}(K_t^b)^{\top} \end{bmatrix}, \hspace{5mm} \hat{B}_{u, t} =  \begin{bmatrix} B_{u, t} \\ 0 \end{bmatrix}, $$
$$\hat{B}_{w, t} =  \begin{bmatrix} B_{w, t} (\Sigma_t^b)^{-1/2} \\ B_{w, t} (\Sigma_t^b)^{-1/2} \end{bmatrix}, \hspace{5mm} \hat{Q}_t = \begin{bmatrix} Q_t & 0 \\ 0 & 0 \end{bmatrix},$$
and $$\hat{H}_t = I + \hat{B}_{u, t}^{\top} \hat{P}_{t+1} \hat{B}_{u, t}, $$ 
and $\hat{P}_t$ is the solution of the backwards Riccati recursion $$\hat{P}_{t} = \hat{Q}_t + \hat{A}_t^{\top}\hat{P}_{t+1}A_t - \hat{A}_t^{\top} \hat{P}_{t+1} \hat{B}_{u, t} \hat{H}_t^{-1} \hat{B}_{u, t}^{\top} \hat{P}_{t+1} \hat{A}_t $$ where we initialize $\hat{P}_T = 0$.
 We emphasize that the driving disturbance in this system is not $w$, but rather $w' = \Delta_2 w$. In (\ref{model-for-delta2}) we found a state-space model for $\Delta$, so it is easy to see that a state-space model for $w'$ is given by 
  $$\nu_{t+1} =  \tilde{A}_t\nu_t + B_{w, t} w_t, $$
 $$ w_t' = (\Sigma_t^b)^{1/2} (K_t^b)^{\top}\nu_t  +   (\Sigma_t^b)^{1/2}w_t, $$
where we initialize $\nu_0 = 0$. Substituting this expression for $w'_t$ in place of $z_t$ in (\ref{regret-optimal-controller-description-proof}) and simplifying, we obtain the state-space description of the regret-optimal causal controller in Theorem \ref{regret-optimal-controller-thm}. We note that we can easily derive an analogous state-space description of the regret-optimal strictly causal controller simply by plugging the dynamics (\ref{regret-final-dynamics}) into the state-space model for the strictly causal $H_{\infty}$ controller (see Theorem 9.5.2 in \cite{hassibi1999indefinite} for such a model). .
The regret bound stated in Theorem \ref{regret-optimal-controller-thm} are immediate: by definition the regret-suboptimal controller at performance level $\gamma$ has dynamic regret at most $\gamma^2 \|w\|_2^2$. To find the optimal value of $\gamma$, we minimize $\gamma$ subject to the constraints described in Theorem \ref{hinf-suboptimal-controller-thm}.
$\blacksquare$

\section{Integrating predictions and delay} \label{predictions-delay-sec}
In this section, we extend the results of Section \ref{regret-optimal-control-sec} to settings with predictions and delay by using reductions to Theorem \ref{regret-optimal-controller-thm}. While we consider control with predictions and control with delay separately, we emphasize that it easy to extend our results to settings with both predictions and delay by performing one reduction and then the other.

\subsection{Regret-optimal control with predictions}
Consider a system with dynamics given by the linear evolution equation (\ref{evolution-eq}), where the controller can predict future disturbances over a horizon of length $h$; in other words, at time $t$ the controller knows the disturbances $w_{t}, \ldots w_{t+h -1}$.  Define the augmented state $$\xi_t = \begin{bmatrix} x_t \\ w_{t} \\ w_{t+1} \\ \vdots \\ w_{t + h -2} \\ w_{t+h -1} \end{bmatrix}. $$ We can think of $\xi_t$ as representing the state $x_t$ along with a transcript of the next $h$ predicted disturbances $w_{t}, \ldots, u_{w+h-1}$. Notice that $\xi$ has dynamics given by the linear evolution equation 

\begin{equation} \label{transformed-pred-eq}
\xi_{t+1} = \hat{A}_t \xi_t + \hat{B}_{u, t}u_t + \hat{B}_{w, t}w_t',
\end{equation}
where we define $w_t' = w_{t+h}$ and
\begin{align*}
\hat{A}_t = \begin{bmatrix} A_t & B_{w, t}  & 0 & \ldots & 0 & 0 \\ 0 & 0 & I & \ldots & 0 & 0 \\ 0  & 0 &  &  &  & 0 \\ \vdots & \vdots  &  & \ddots  & \ddots &  \\ \vdots & \vdots  & & &  & I \\ 0 & 0 & 0 & \ldots & 0 & 0 \end{bmatrix}, \hspace{5mm}  \hat{B}_{u, t} = \begin{bmatrix} B_{u, t} \\ 0 \\ 0 \\  \vdots \\ 0 \\ 0 \end{bmatrix} u_t, \hspace{5mm} \hat{B}_{w, t} = \begin{bmatrix} 0 \\ 0 \\ 0 \\ \vdots \\ 0 \\ I \end{bmatrix}.
\end{align*} 
We can rewrite the LQR cost  in the original system (\ref{evolution-eq}) in terms of $\xi$: 
\begin{equation} \label{new-lqr-cost-predictions}
x_T^{\top}Q_T x_t + \sum_{t=0}^{T-1} \left( x_t^{\top}Q_t x_t + u_t^{\top}R_t u_t \right) = \xi_T^{\top}\hat{Q}_T \xi_t + \sum_{t=0}^{T-1} \left( \xi_t^{\top}\hat{Q}_t \xi_t + u_t^{\top}R_t u_t \right),
\end{equation}
where we define the block-diagonal matrix $$\hat{Q}_t = \begin{bmatrix} Q & 0 & \ldots \\ 0 & 0 \\ \vdots & & \ddots \end{bmatrix}.$$ In light of (\ref{new-lqr-cost-predictions}), we see that any sequence of control actions $u = (u_0, \ldots, u_{T-1})$ generates an identical cost in the original system (\ref{delay-eq}) and in the new system (\ref{transformed-delay-eq}). Notice that a causal control policy in the new system (i.e. one whose control actions at time $t$ depend only on $w_0', \ldots w_t')$ is a policy with a lookahead of length $h$ in the original system. We have proven: 
 
\begin{theorem} \label{regret-optimal-mpc-theorem}
The regret-optimal model-predictive controller with lookahead of length $h$ in the dynamical system (\ref{evolution-eq}) is the regret-optimal controller described in Theorem \ref{regret-optimal-controller-thm} applied to the system (\ref{transformed-pred-eq}). %Similarly, the causal controller which minimizes competitive ratio in the system (\ref{pred-eq}) is the controller with optimal competitive ratio described in Theorem (\ref{cr-thm}) in the system (\ref{transformed-pred-eq}).
 \end{theorem}

\subsection{Regret-optimal control with delay}
Consider the linear evolution equation 

\begin{equation} \label{delay-eq}
x_{t+1} = A_t x_t + B_{u, t-d}u_{t-d} + B_{w, t} w_t.
\end{equation}
In this system, control actions affect the state only after a delay of length $d$; in other words, at time $t$, the state $x_t$ is a function of $w_0, w_1, \ldots w_{t-1}$ and $u_0, u_1, \ldots u_{t-d-1}$. For $t = 0, \ldots T-1$, we define the augmented state $$\xi_t = \begin{bmatrix} x_t \\ u_{t-1} \\ u_{t-2} \\ \vdots \\ u_{t-d + 1} \\ u_{t-d} \end{bmatrix}. $$ We can think of each $\xi_t$ as representing the actual state $x_t$ along with a transcript of the previous $d$ control actions. Notice that $\xi$ has dynamics given by the linear evolution equation 
\begin{equation} \label{transformed-delay-eq}
\xi_{t+1} = \hat{A}_t \xi_t + \hat{B}_{u, t} u_t + \hat{B}_{w, t}w_t,
\end{equation}
where we define 
\begin{align*}
\hat{A}_t =  \begin{bmatrix} A_t & 0  & 0 & \ldots & 0 & B_{u, t - d} \\ 0 & 0 & 0 & \ldots & 0 & 0 \\ 0  & I &  0 & \ldots & 0 & 0 \\ \vdots &  &  \ddots & &  & \vdots \\ 0 &  & & \ddots &  & \vdots \\ 0 & 0 & 0 & \ldots & I & 0 \end{bmatrix}, \hspace{5mm} \hat{B}_{u, t} =  \begin{bmatrix} 0 \\ I \\ 0 \\  \vdots \\ 0 \\ 0 \end{bmatrix}, \hspace{5mm} \hat{B}_{w,t} =  \begin{bmatrix} B_{w, t} \\ 0 \\ 0 \\ \vdots \\ 0 \\ 0 \end{bmatrix}.
\end{align*}
We can rewrite the LQR cost in the original system (\ref{delay-eq}) in terms of $\xi$: 
\begin{equation} \label{new-lqr-cost-delay}
\sum_{t=0}^{T-1} \left( x_t^{\top}Q_t x_t + u_t^{\top}R_t u_t \right) = \xi_T^{\top}\hat{Q}_T \xi_t + \sum_{t=0}^{T-1} \left( \xi_t^{\top}\hat{Q}_t \xi_t + u_t^{\top}R_t u_t \right),
\end{equation}
where we define the block-diagonal matrix $$\hat{Q}_t = \begin{bmatrix} Q & 0 & \ldots \\ 0 & 0 \\ \vdots & & \ddots \end{bmatrix}.$$ In light of (\ref{new-lqr-cost-delay}), we see that any sequence of control actions $u = (u_0, \ldots, u_{T-1})$ generates an identical cost in the original system (\ref{delay-eq}) and in the new system (\ref{transformed-delay-eq}). Also notice that the new system is of the form (\ref{evolution-eq}), i.e. it has no delay. A causal controller in the new system is a causal controller in the original system. We have proven: 
 
\begin{theorem} \label{regret-optimal-delay-theorem}
The regret-optimal controller in the dynamical system with delay (\ref{delay-eq}) is the regret-optimal controller described in Theorem \ref{regret-optimal-controller-thm} applied to the system (\ref{transformed-delay-eq}). 
\end{theorem}

\end{document}